\documentclass[10pt,a4paper,twocolumn,fleqn]{article}

\usepackage[T1]{fontenc}
\usepackage[utf8]{inputenc}
\usepackage{amsmath}
\usepackage{amsthm}
\usepackage{newtxtext,newtxmath}   
\usepackage{microtype}
\allowdisplaybreaks

\usepackage[a4paper, hmargin=18mm, vmargin={22mm,22mm}, columnsep=18pt]{geometry}

\usepackage{graphicx}
\usepackage{xcolor}
\usepackage{booktabs}
\usepackage{array}
\usepackage{siunitx}
\usepackage{nicefrac}
\usepackage{placeins}
\usepackage{float}
\usepackage{stfloats}   
\usepackage{afterpage}
\usepackage{etoolbox}
\usepackage[export]{adjustbox}

\usepackage{tikz}
\usetikzlibrary{arrows.meta, positioning, fit, calc, angles, quotes}
\pgfdeclarelayer{background} 
\pgfsetlayers{background,main} 

\usepackage[font=small,labelfont=bf]{caption}
\usepackage{subcaption}

\usepackage[authoryear]{natbib}

\usepackage{hyperref}
\usepackage{orcidlink}

\makeatletter
\renewcommand\section{\@startsection
  {section} 
  {1}       
  {0pt}     
  {3.0ex plus 1ex minus .2ex} 
  {2.0ex plus .2ex}           
  {\normalfont\normalsize\bfseries}}
\renewcommand\subsection{\@startsection
  {subsection} 
  {2}          
  {0pt}        
  {3.0ex plus 1ex minus .2ex} 
  {2.0ex plus .2ex}           
  {\normalfont\normalsize\itshape}}
\renewcommand\subsubsection{\@startsection
  {subsubsection} 
  {3}             
  {0pt}           
  {2.0ex plus 1ex minus .2ex} 
  {0.2ex plus .2ex}           
  {\normalfont\normalsize\itshape}}
\makeatother

\definecolor{AccentBlue}{RGB}{0,166,214}
\hypersetup{
  colorlinks=true,
  linkcolor=AccentBlue,
  citecolor=AccentBlue,
  urlcolor=AccentBlue,
  pdftitle={Mechanical Precision Weeding with a Quadruped Robot},
  pdfauthor={Ruben Beumer, Tom Janssen, René van de Molengraft, Duarte Antunes}
}

\theoremstyle{definition}
\newtheorem{contr}{Contribution}

\newcolumntype{L}{@{\extracolsep{\fill}}l}
\newcolumntype{R}{@{\extracolsep{\fill}}r}
\newcolumntype{C}{@{\extracolsep{\fill}}c}
\newcommand{\tblwidth}{.95\linewidth}
\AtBeginEnvironment{table}{\centering}

\renewenvironment{abstract}
  {\par\medskip\noindent{\bfseries Abstract}\par\smallskip\noindent\ignorespaces}
  {\par}
\newenvironment{keywords}
  {\par\medskip\noindent{\bfseries Keywords:} \ignorespaces}
  {\par}
\newcommand{\sep}{\unskip,\ }

\begin{document}

\makeatletter
\twocolumn[{%
\begin{@twocolumnfalse}
  \begin{center}
    {\LARGE\bfseries Mechanical Precision Weeding with a Quadruped Robot\par}
    \vspace{1.2em}
    {\large
      Ruben Beumer\,\orcidlink{0009-0009-5261-4374}\textsuperscript{\,a,*},\quad
      Tom Janssen\,\orcidlink{0009-0004-2108-2081}\textsuperscript{\,a,\textdagger},\quad
      René van de Molengraft\,\orcidlink{0000-0002-5095-4297}\textsuperscript{\,a},\quad
      Duarte Antunes\,\orcidlink{0000-0003-3047-9334}\textsuperscript{\,a}\par}
    \vspace{0.8em}
    {\small\itshape
      \textsuperscript{a}Department of Mechanical Engineering, Eindhoven University of Technology,
      P.O. Box 513, 5600 MB Eindhoven, The Netherlands\par}
  \end{center}
  \vspace{0.5em}
  \noindent\rule{\textwidth}{0.4pt}
  %
  \begin{abstract}
    Herbicide-based weed control is increasingly unsustainable due to rising weed resistance and the adverse environmental impacts of chemical use. While mechanical weed control avoids these drawbacks, it is typically implemented using large machines that cause soil compaction. We propose a novel alternative based on small mobile robots for mechanical weeding. Compared with existing automated mechanical weeding approaches, the proposed method offers reduced soil compaction, simpler automation, and improved scalability. Our solution involves a Boston Dynamics Spot quadruped robot equipped with a custom weed removal tool featuring a milling bit at its end.
    The tool is rigidly attached to the robot and uses the degrees of freedom of the robot base by actuating the legs, while keeping the feet stationary.
    We develop a software architecture that enables autonomous weed removal and integrate this system with all other required components.
    We analyze the accuracy and efficiency of the current proof of concept both in an indoor and outdoor environment and provide recommendations for future work to make the system more accurate and efficient.
    
\end{abstract}

  %
  \begin{keywords}
  Mechanical weeding \sep Precision weeding \sep Robot dog \sep Quadruped
  \end{keywords}
  \smallskip
  \noindent\rule{\textwidth}{0.4pt}
  \vspace{1.5em}
\end{@twocolumnfalse}
}]
\makeatother

\begingroup
\renewcommand{\thefootnote}{\fnsymbol{footnote}}
\footnotetext[1]{Corresponding author.\\
\textit{E-mail addresses}:
\texttt{\href{mailto:r.m.beumer@tue.nl}{r.m.beumer@tue.nl}} ({R.\,M.} Beumer),
\texttt{\href{mailto:tjanssen@ethz.ch}{tjanssen@ethz.ch}} ({T.\,M.\,E.} Janssen),
\texttt{\href{mailto:m.j.g.v.d.molengraft@tue.nl}{m.j.g.v.d.molengraft@tue.nl}} ({M.\,J.\,G.} v.d. Molengraft),
\texttt{\href{mailto:d.antunes@tue.nl}{d.antunes@tue.nl}} ({D.} Antunes)}
\footnotetext[2]{Present address: Institute for Transport Planning and Systems, ETH Zürich, Stefano-Franscini-Platz 5, Zürich, Switzerland.}
\endgroup

\section{Introduction}
\label{sec:introduction}

Weeds pose a major challenge in agriculture, as they compete with crops for limited resources like water, nutrients, space, and sunlight. Their aggressive, adaptive and persistent nature makes them especially threatening to crop production during the early growth stages before canopy closure, when young crops are most vulnerable to being outcompeted, ultimately reducing agricultural yield \citep{Kaur2018, Zimdahl2004}.
Consequently, farmers employ a range of weed management techniques to mitigate the impact of these undesired plants. Chemical weed management, i.e., spraying herbicides, is the most effective and economical way to manage weeds, and the most widely used \citep{Manisankar2022, Harker2013}.

However, resistance of weeds to popular herbicides continues to increase \citep{Heap2024}, requiring more research on alternative herbicides in the future \citep{Shaner2014}. Moreover, using herbicides alone for weed management is not sustainable, as herbicide resistance can only be managed and never eliminated \citep{Shaner2014b}.
In addition to the problem of herbicide resistance, herbicides pose a risk to both human and environmental health. For instance, glyphosate-based herbicides have been associated with liver and kidney damage, endocrine disruption, DNA damage, reproductive problems, and various neurodegenerative disorders \citep{Thongprakaisang2013, Seralini2014, Kwiatkowska2017}.

Clearly, there is a need to drastically reduce or eliminate the use of herbicide-assisted weed removal. This requires alternative weed removal strategies.

\subsection{Alternative weed removal techniques}
\label{sec:alternative}
Currently, there are numerous weed removal systems that minimize or eliminate herbicide use. Pre-planting tillage is a common practice employed to remove existing weeds in the field by burying them deeper in the soil profile, thereby reducing their potential for emergence. After planting, but before the crop emerges, blind cultivation can be used to dislodge weeds whose root systems are shallower than those of the crop, providing early weed control. Once the weeds have emerged, various mechanical tools are employed for inter-row cultivation, effectively managing weeds between crop rows but not within them. Intra-row weeding presents a greater challenge; nevertheless, a wide variety of tools are available or being developed to remove these weeds \citep{Das2025}. These tools often require the crop to be more established than the weeds, as they disrupt the entire crop row, dislodging anything insufficiently rooted, or they depend on how accurate a system can distinguish between weeds and crops. 

In recent years, several (semi-)autonomous robotic solutions have been developed that are capable of performing both inter-row and intra-row weed removal, irrespective of the crop-weed condition. These systems generally utilize cameras for the detection and differentiation of weeds from the desired crop.
\citeauthor{Trabotyx} developed a robot capable of autonomously navigating through field rows to remove weeds using a spinning milling bit. It requires operator assistance for turning to move from one row to another, thus it cannot operate completely autonomously.
\citeauthor{Ecorobotix} developed a 6-meter-wide tractor attachment that selectively sprays herbicides at weed locations, reducing herbicide usage by up to 95\%.
\citeauthor{Carbonrobotics} developed a similar attachment to that of Ecorobotix, but uses lasers instead of chemicals to remove weeds.
\citeauthor{RobotOne} developed a robot that is tele-operated to navigate the field, while autonomously removing weeds using laser burning. 
\citeauthor{OddBot} developed an autonomous robot that removes weeds by picking and pulling them using a small gripper.

\subsection{Design requirements: small mobile robots as an alternative}
\label{sec:requirements}
The systems listed in \autoref{sec:alternative} and other related solutions typically rely on tractor attachments or machines with a weight comparable to that of a tractor.
Our objective is to design a system that offers several significant advantages over traditional practices, not only for weed removal tasks, but in agricultural automation in general: it should (1) reduce soil compaction, (2) be scalable, and (3) provide easy integration with intercropping. 
Hence, we propose a radically different approach: utilizing a swarm of small, lightweight, fully autonomous mobile robots. Full autonomy is becoming increasingly important given the growing labor shortages in the agricultural sector. Furthermore, when operating in proximity to humans, animals, and objects, small autonomous mobile robots are safer than larger systems due to their lower weight, which reduces potential damage in control failures and could simplify regulatory approval for fully autonomous operations. In the following, we explain how deploying such a swarm of small robots achieves the desired advantages.

\textbf{Reduced soil compaction}: Heavy machinery significantly compacts the soil, limiting water infiltration and root growth, resulting in lower crop yields \citep{Lipiec2003}. In contrast, mobile robots, which can be an order of magnitude lighter than traditional tractors, can substantially mitigate this issue.

\textbf{Scalability}: Although the smaller size and reduced payload capacity of mobile robots mean they perform less work in a given time compared to larger machines, this limitation can be overcome by deploying multiple robots to collaboratively complete tasks. This modular approach offers flexible scalability (as smaller units offer greater flexibility in matching capacity to need), reducing inefficiencies, and limiting productivity loss if a single robot fails, unlike relying on one large system.

\textbf{Easy integration with intercropping:} Intercropping is the cultivation of different crops within the same field in a structured or unstructured way. It offers several advantages over large mono-cropped fields. Diverse crops can serve as natural barriers to diseases \citep{Boudreau2013}, enhance soil and insect biodiversity, leading to an increase in available nutrients, and the density of crops in a field can be increased, by, e.g., placing crops with deep roots near crops with shallow roots \citep{Cobbenhagen2021}. Intercropping has been shown to increase the yield by 22\% on average compared to mono-cropping \citep{Yu2015}. Despite these benefits, large-scale adoption of intercropping remains limited due to the inadequacy of conventional machinery for such practices and the high cost of manual labor. The deployment of a swarm of small mobile robots allows for easier adaptation of intercropping as every robot can differ slightly to accommodate for required differences between crops.

The system requires three main functions: (1) the ability to move through the field (\textbf{locomotion}) and, for the precision weeding task itself, (2) weed \textbf{detection} and (3) weed \textbf{removal}. As our focus is on mechanical feasibility and not on the detection of weeds, and given the existence of various successful systems employing similar detection methods, as discussed in \autoref{sec:alternative}, we assume that the detection of weeds using a RGB-D camera will achieve sufficient accuracy to allow testing the other components of the system.

\subsection{Mobile robot selection}
Selecting a mobile robot for an agricultural application depends on two main factors: its locomotion system, which must enable it to navigate the field effectively, and its ability to carry and operate a task-specific tool. Various locomotion methods, with their respective advantages and disadvantages for weed removal, are summarized below.

\textbf{Wheeled locomotion} is energy-efficient, mechanically simple, and relatively inexpensive to manufacture and maintain. However, wheeled robots have limited adaptability to rough, muddy, or uneven terrain, requiring well-prepared fields. In addition, due to the (usually) fixed position and dimensions of the wheels, row spacing must allow passage, and sufficient space is needed for turning and entering subsequent rows.

\textbf{Tracked locomotion} uses continuous tracks, offering better terrain traversal due to their ability to distribute weight more evenly. These systems are more complex and more expensive to manufacture and maintain, and fields still require adequate row spacing and turning space.

In recent years, \textbf{legged locomotion} has gained popularity. Legged robots are able to navigate flat, uneven, or muddy fields without relying on row spacing. They can move freely around crops and between rows without dedicated turning space, making them especially useful during early growth stages when they can step over a row of crops to a different row. However, they are generally less energy-efficient, more expensive, and more difficult to maintain than their wheeled counterparts.

Finally, \textbf{flying mobile robots (drones)} are also available. However, due to the limited flight time and the force requirements associated with interaction tasks such as mechanical weed removal, we did not consider these systems for this application.

For a generally applicable weed removal system, robust locomotion in various field conditions is a critical requirement. Additionally, the robot must be adaptable to different row configurations within the field to allow for wide deployability. Consequently, we selected legged robots as the most suitable type of locomotion.

In legged locomotion, two main configurations are employed: quadruped robots and bipedal robots, the former being much more robust and stable. Quadrupeds have been developed that are able to complete difficult routes under a variety of conditions \citep{Hoeller2024, Lee2020}, whereas state-of-the-art humanoid (bipedal) robots are more limited \citep{He2024, Haarnoja2024}, and have not yet been shown to work robustly outside of a lab setting. In addition, quadrupeds might be the more suitable choice even when bipedals become more robust, as quadruped platforms are by definition more stable because they have four legs, and can more naturally operate close to the ground. 

Quadrupeds that have been developed to interact with their environment typically employ either their legs for relatively simple tasks \citep{Schwarke2023} or a multi-degree-of-freedom robotic arm for more complex tasks, effectively decoupling locomotion from task handling \citep{Ferrolho2023, Bellicoso2019, Zimmermann}. Quadrupeds typically have a payload capacity high enough to carry such a tool to interact with the environment, which makes them suitable for our application.

For the task of weed removal, robotic arms that decouple manipulation from locomotion could be employed, but this adds extra complexity, cost and weight. Instead, the degrees of freedom of the robot's body available when the robot is not walking can be used to command a 0-degree-of-freedom (0-DOF) robotic arm with a 1-DOF end-effector for milling. This simpler design is expected to be sufficient for the task while reducing system complexity.

\subsection{Contributions}
This work presents the development and feasibility testing of a quadruped robot for mechanical weed removal. The system employs a quadruped platform with a custom 0-DOF robotic arm on top of it with an end-effector to mill. This allows for inter-row as well as intra-row weed removal in a variety of field conditions for a large variety of crops in the pre-canopy-closure period, eliminating the need for herbicides. To this end, the following contributions are made:

\begin{contr}
\label{contr:1}
We select components and integrate a hardware architecture including all components for mechanical weed removal by a quadruped robot with a custom-built 0-DOF robotic arm.
\end{contr}

\begin{contr}
\label{contr:2}
We develop a complete control pipeline that encompasses all the required software to make our custom setup remove weeds. This gives insight into the analysis of critical components and can be used as a starting point for further, more component-specific, studies.
\end{contr}

\begin{contr}
\label{contr:3}
We analyze the technical feasibility of using quadrupeds equipped with a manipulation tool in an agricultural setting, conducting tests both indoors and outdoors across various terrains, including flat, uneven, soft and rigid ground, giving insights into performance in actual agricultural settings.
\end{contr}

\begin{contr}
\label{contr:4}
We analyze the efficiency of our setup and compare it with the state-of-the-art robotic solutions that are currently available for mechanical weed removal, to conclude if our system is or can become a realistic solution for the large-scale requirements of the agricultural sector.
\end{contr}

The remainder of this paper is structured as follows. \autoref{sec:design} discusses the design and integration. \autoref{sec:control-architecture} continues on this setup and elaborates on the control pipeline made specifically for the previously introduced design. In \autoref{sec:results}, the system is tested and results related to accuracy and efficiency are reported. These results and the contributions made are discussed in \autoref{sec:conclusions}. Lastly, further improvements and extensions are recommended in \autoref{sec:recommendations}. The developments and results are also demonstrated in a video\footnote{\url{https://www.youtube.com/watch?v=KtmNKJLgfKo}}.

\section{Design and integration}
\label{sec:design}

\subsection{Quadruped platform}

\begin{figure}
    \centering
    \includegraphics[height=4.1cm, trim={15cm 12cm 15cm 16cm}, clip, valign=b]{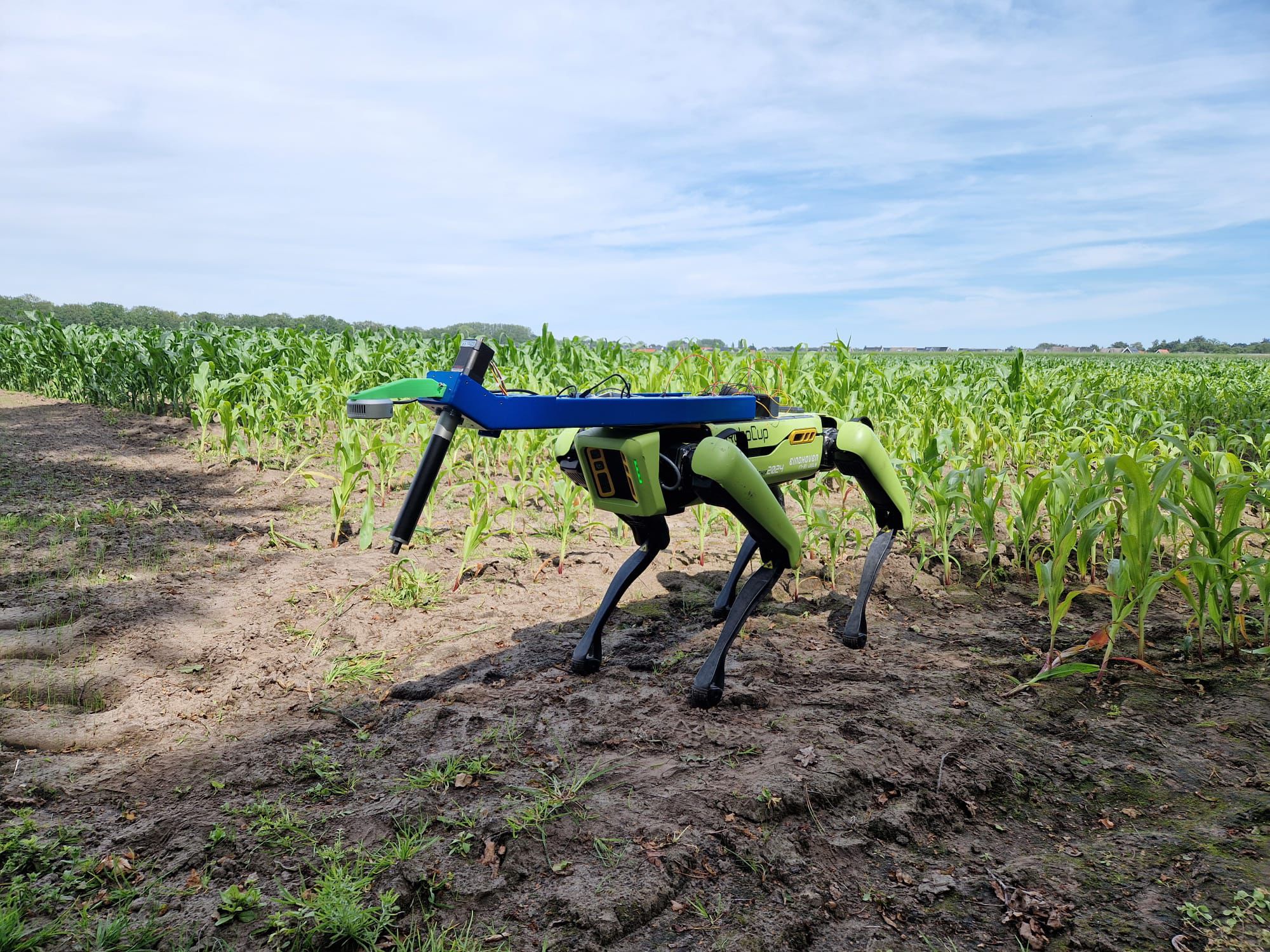}%
    \hfill
    \includegraphics[width=4.1cm, angle=-90, valign=b, trim={10cm 29cm 8cm 29cm}, clip]{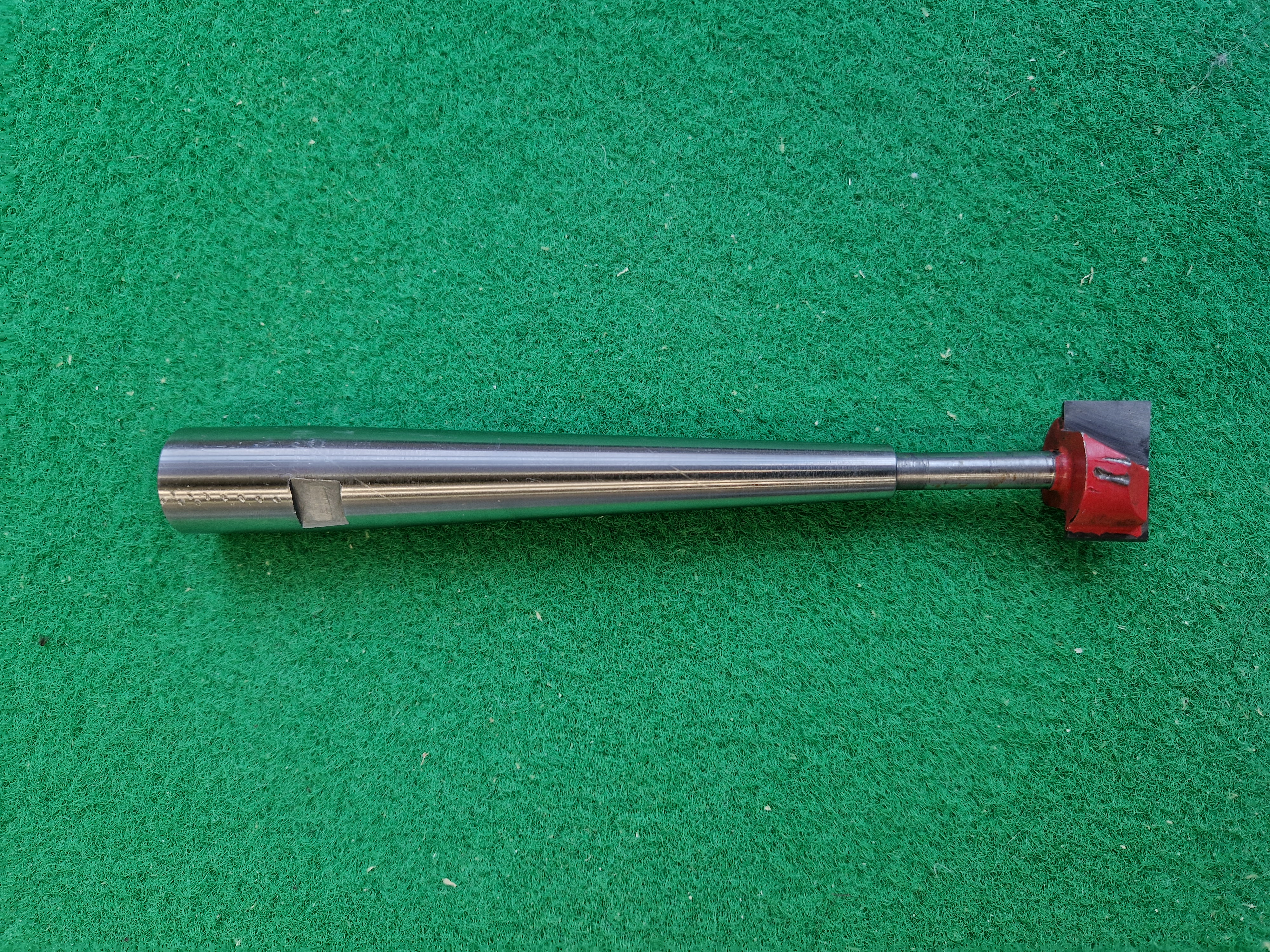}%
    \caption{Left: Spot equipped with a custom weeding tool in a maize field. Right: the end-effector.}
    \label{fig:spot-with-tool}
\end{figure}

The \citeauthor{bdspot} Spot robot (see \autoref{fig:spot-with-tool}) serves as the quadruped platform for this study. Spot is suitable because of its robust control capabilities and size, as it is large enough to support a tool while compact enough to maneuver between plants. It offers resilient and easily deployable locomotion capabilities, as well as the ability to perform both rotational and translational movements while keeping its feet in the same location, thereby meeting all necessary operational requirements. It should be noted that other legged robots with similar characteristics, such as \citeauthor{anybotics}'s ANYmal or \citeauthor{Unitree}'s B2, could also have been employed. The methodology proposed in this work is designed to be interoperable and can also be applied to these alternative platforms.

The robust capabilities of Spot come with the drawback of restricted access to the low-level code. Instead, interaction with Spot is facilitated through a Python API, which supports high-level commands. This API allows users to issue commands related to, e.g., locomotion, utilizing predefined gaits, adjusting the robot's pose, and retrieving the robot's states such as joint angles and sensor data.

\begin{figure}
    \centering
    \includegraphics[width=0.85\linewidth]{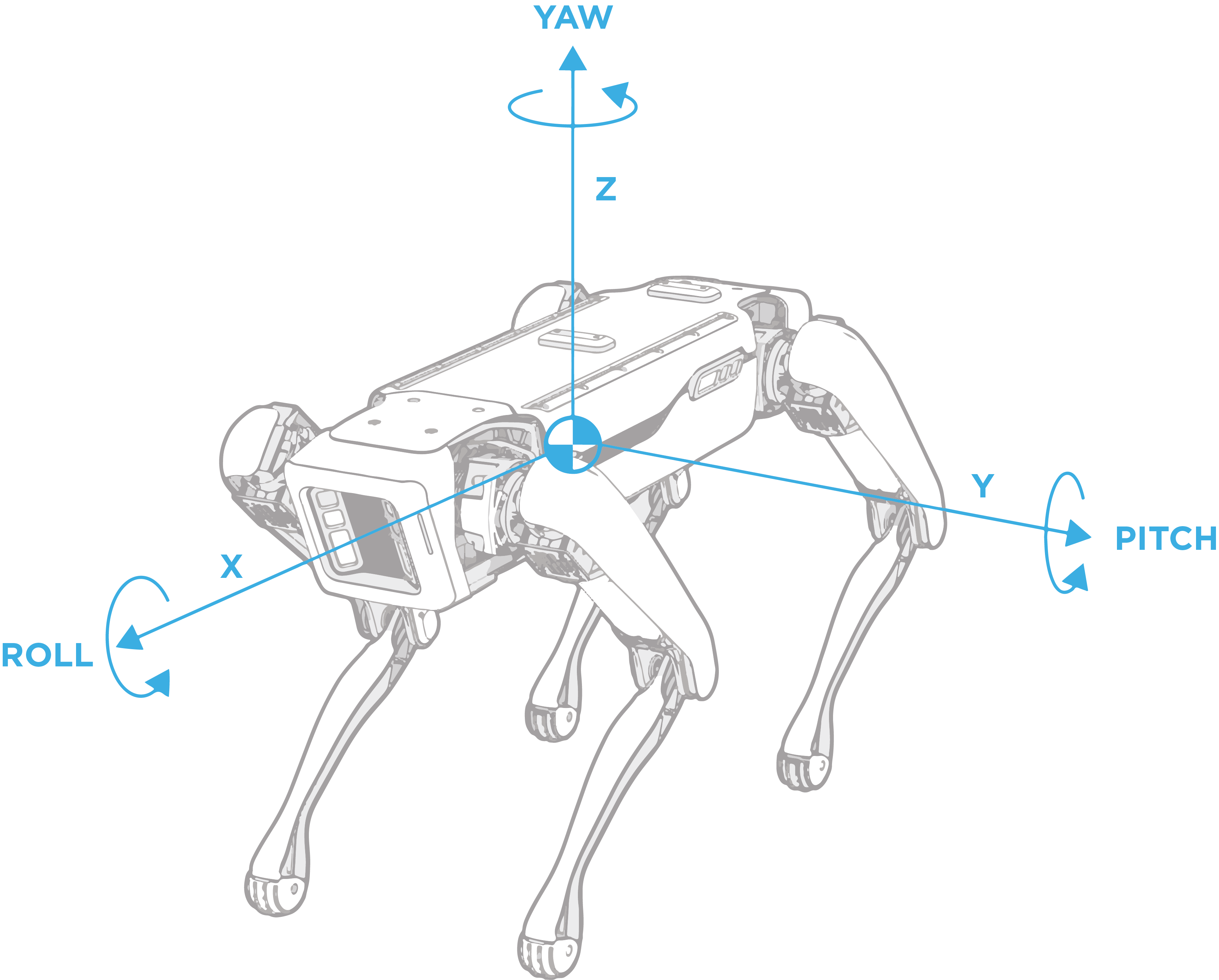}
    \caption{The body frame of Spot (\citeauthor{bdspot}).}
    \label{fig:spot-body-frame}
\end{figure}

The specific API functionality used for weed removal using the custom tool is the body pose command that receives the desired translation and rotation of the body frame (see \autoref{fig:spot-body-frame}) as input. This pose reference is tracked using Spot's internal controllers, having a limited accuracy and precision. These values, relevant to determine whether weed removal using the custom tool is feasible, are experimentally found using 75 different body poses and reported in \autoref{tab:pose-accuracy}. These values are found using Spot's internal state estimator, which is based on the joint angles and kinematics. This state estimator is validated first using a custom kinematic analysis based on the dimensions of Spot and the joint readings that are assumed to be accurate, given that they are read out by high-precision joint angle encoders. As the end-effector milling bit responsible for destroying the weeds is \SI{22}{\milli\meter} in diameter (see \autoref{sec:tool}), its center can only deviate about \SI{11}{\milli\meter} from the target weed in any direction in order to hit the center of its stem.
The difference between the theoretical end-effector position based on the command to move to \mbox{$x=\SI{0.08}{\meter}$}, \mbox{$\theta=\SI[parse-numbers=false]{\nicefrac{\pi}{10}}{\radian}$} and \mbox{$\psi=\SI[parse-numbers=false]{\nicefrac{\pi}{10}}{\radian}$} following from the forward kinematics (see \autoref{sec:kinematics}), and the theoretical end-effector position based on the same command including bias (i.e., when adding the mean errors from \autoref{tab:pose-accuracy} in the kinematics), is reported in \autoref{tab:resulting-deviation} and falls within the requirements. The standard deviation of the error in $y$-direction is relatively large. However, a large movement in a certain direction followed by a small movement results in both movements having approximately the same error. This means that a large error can also be compensated using a feedback control correction step, as further explained in \autoref{sec:removal}.

\begin{table}[ht]
    \caption{The mean error $\mu$ and standard deviation $\sigma$ for each Degree Of Freedom (DOF) in their respective units.}
    \label{tab:pose-accuracy}
    \centering
    \makebox[0.95\linewidth]{%
        \begin{minipage}{0.38\linewidth}
            \centering
            \begin{tabular*}{\linewidth}{@{} LRR@{} }
                \toprule
                DOF & $\mu$ [mm] & $\sigma$ [mm]\\
                \midrule
                $x$ & 1.11 & 6.33 \\
                $y$ & 4.79 & 27.88 \\
                $z$ & 0.33 & 2.04 \\
                \bottomrule
            \end{tabular*}
        \end{minipage}
        \hfill
        \begin{minipage}{0.52\linewidth}
            \centering
            \begin{tabular*}{\linewidth}{@{} LRR@{} }
                \toprule
                DOF & $\mu$ [$10^{-3}~\!\text{rad}$] & $\sigma$ [$10^{-3}~\!\text{rad}$]\\
                \midrule
                $\phi$ & 0.84 & 4.23 \\
                $\theta$ & 3.34 & 8.14 \\
                $\psi$ & -1.78 & 2.23 \\
                \bottomrule
            \end{tabular*}
        \end{minipage}
    }
\end{table}

\begin{table}[ht]
    \caption{The position of the end-effector in the world frame when Spot is commanded to move to \mbox{$x=\SI{0.08}{\meter}$}, \mbox{$\theta=\SI[parse-numbers=false]{\nicefrac{\pi}{10}}{\radian}$}, \mbox{$\psi=\SI[parse-numbers=false]{\nicefrac{\pi}{10}}{\radian}$}, based on the forward kinematics, with and without bias from \autoref{tab:pose-accuracy}, and the resulting error.}
    \label{tab:resulting-deviation}
    \begin{tabular*}{\tblwidth}{@{} LRRR@{} }
        \toprule
            DOF & Without bias [mm]& With bias [mm]& Error [mm]\\
        \midrule
            $x_{\text{EF}}$ &  715.44 &  715.14 &  0.29 \\
            $y_{\text{EF}}$ &  206.47 &  209.85 & -3.39 \\
            $z_{\text{EF}}$ &  -32.53 &  -34.41 &  1.88 \\
        \bottomrule
    \end{tabular*}
\end{table}

Additionally, the API supports payload configuration, which involves specifying parameters such as mass, center of mass, and moment of inertia. This payload information is critical for the internal control system to adjust for variations during movement. Spot can carry a maximum payload capacity of \SI{14}{\kilo\gram}. Communication with Spot can be established via LAN, through a Wi-Fi network, or by utilizing a local network set up by Spot itself. For remote field operations, Spot's self-established network is employed to facilitate communication.

\subsection{Weeding tool}
\label{sec:tool}
A custom tool developed by \citeauthor{BromMechatronica} is employed for weed removal, mounted on top of Spot to its mounting rails, featuring a milling bit that is \SI{22}{\milli\meter} in diameter protruding at the front, as illustrated in \autoref{fig:spot-with-tool}. The milling bit is connected via an extended shaft to a NEMA 17 stepper motor, which provides the necessary torque for effective operation. This motor is controlled by a Nanotec C5-E controller that requires \SI{48}{\volt}. Given that Spot's power output ranges from \SI{35}{\volt} to \SI{59}{\volt}, depending on battery level, a DC-DC converter is incorporated to ensure the controller consistently receives the correct voltage. The Nanotec controller facilitates precise speed control of the end-effector up to 7000 rpm, providing sufficient torque and speed for efficient weed removal.

The tool weighs approximately \SI{7}{\kilo\gram} and is engineered to withstand the vibrations induced by the motor. The weight is centered approximately \SI{25}{\centi\meter} from the base of the tool extending in front of Spot's front legs, resulting in a forward shift in Spot's center of mass. This forward weight distribution is compensated for by configuring a payload in Spot's software, ensuring stable and balanced operation.

\subsection{Camera}
\label{sec:camera-setup}
A camera is used to perceive the environment and detect weeds. Spot already has five built-in grayscale stereo vision cameras that are used for its localization, locomotion, and obstacle avoidance. However, due to the low resolution and the grayscale, plants and weeds can hardly be detected by these cameras, and thus a different camera is required.

To determine weed locations, we placed an Intel RealSense L515 camera on top of the weed removal tool using a custom 3D-printed mount. This device contains an RGB camera as well as a LiDAR depth camera, suitable for applications that require high precision and high resolution, which allows 3D point extraction of weeds visible by the camera. The camera is placed slightly more than \SI{25}{\centi\meter} above the end-effector to achieve accurate depth readings, as this is the camera's minimum depth range.

\subsection{Integration}
A schematic overview of the hardware architecture is shown in \autoref{fig:integration}. In \autoref{fig:integration-rpi}, a Raspberry Pi 4B, powered by an external battery, is placed on the tool as a system with which the tool controller and camera can interface. This Raspberry Pi transmits the desired end-effector rotational velocity commands to the tool's controller and receives image data from the camera. As the Raspberry Pi 4B is not powerful enough to run the weed detection algorithm directly (see \autoref{sec:weed-detection}), it communicates with an external PC over Spot's local network using TCP to ensure a stable connection. This PC serves as the main computational unit, running the complete control architecture (see \autoref{sec:control-architecture}). Thus, the PC connects to Spot over Spot's network to send desired pose commands. The required RGB-D images are sent to the PC on its request. When required, the PC sends a desired velocity command for the end-effector to the Raspberry Pi.

Although this control architecture functions effectively, a possible bottleneck for future consideration is the communication latency introduced by the network connection, which complicates time-synchronized control. Therefore, we also implemented a more robust solution in which all computations are performed on a more powerful onboard unit (an NVIDIA Jetson AGX Orin), thereby eliminating the dependency on external processing and minimizing communication delays. This simplified architecture is shown in \autoref{fig:integration-jetson}. Furthermore, the need for an additional external battery was omitted by delivering power to the Jetson from Spot itself using the Spot General Expansion Payload (GXP). For debugging and visualization purposes, a PC can still be wirelessly connected to the Jetson.

\newcommand{\wirelesssymbol}[3][0.25cm]{%
  \fill (#2) circle (1.2pt);
  \draw[wireless, rotate around={{#3+135}:(#2)}] 
        (#2) ++({-#1},0) arc[start angle=180,end angle=270,radius={#1}];
  \draw[wireless, rotate around={{#3+135}:(#2)}] 
        (#2) ++({-0.7*#1},0) arc[start angle=180,end angle=270,radius={0.7*#1}];
  \draw[wireless, rotate around={{#3+135}:(#2)}] 
        (#2) ++({-0.4*#1},0) arc[start angle=180,end angle=270,radius={0.4*#1}];
}

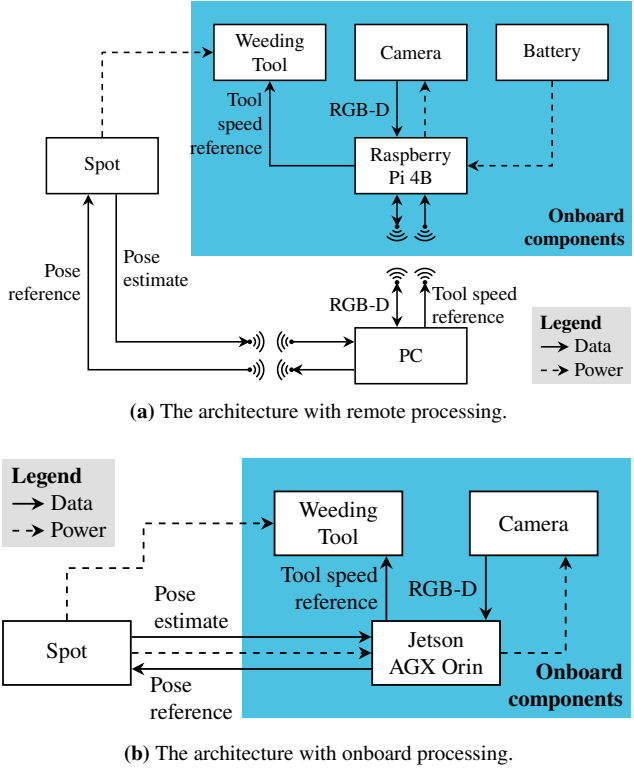
\begin{figure}
    \centering
    
    \begin{subfigure}{\columnwidth}

    \resizebox{\linewidth}{!}{%
    \begin{tikzpicture}[
    box/.style={draw, thick, fill=white, minimum width=2.0cm, minimum height=1.0cm, align=center},
    smallbox/.style={draw, thick, minimum width=2.4cm, minimum height=1.2cm, align=center},
    arrow/.style={thick, -{Stealth[length=2mm,width=2mm]}},
    doublearrow/.style={thick, {Stealth[length=2mm,width=2mm]}-{Stealth[length=2mm,width=2mm]}},
    componentbox/.style={draw=none, fill=AccentBlue!70, inner sep=0.4cm},
    wireless/.style={draw, thick, line cap=round}
    ]

    \node[box] (spot) {Spot};

    \node[box, right=3.5cm of spot] (rpi) {Raspberry\\Pi 4B};

    \node[box, above=1.0cm of rpi] (camera) {Camera};
    \node[box, left=0.5cm of camera] (tool) {Weeding\\Tool};
    \node[box, right=0.5cm of camera] (battery) {Battery};
    
    \node[box, below=2.375cm of rpi] (pc) {PC};
    
    \draw[arrow,dashed] (spot.north) -- ($(spot.north |- tool.west)$) -- node[above]{} (tool.west);
    \draw[arrow,dashed] (battery.south) -- node[right]{} ($(battery.south |- rpi.east)$) -- (rpi.east);
    \draw[arrow,dashed] ($(rpi.north) + (0.25cm,0)$) -- node[right]{} ($(camera.south) + (0.25cm,0)$);
    
    \draw[arrow] (rpi.west) -- ($(tool.south |- rpi.west)$) -- node[left, align=right]{Tool\\speed\\reference} (tool.south);
    
    \draw[arrow] ($(camera.south) + (-0.25cm,0)$) -- node[left]{RGB-D} ($(rpi.north) + (-0.25cm,0)$);
    
    \node[left=1.0cm of pc.west, yshift=0.25cm] (A1) {};
    \node[left=0.5cm of A1] (A2) {};
    \wirelesssymbol{A1}{180}
    \wirelesssymbol{A2}{0}

    \node[left=1.0cm of pc.west, yshift=-0.25cm] (B1) {};
    \node[left=0.5cm of B1] (B2) {};
    \wirelesssymbol{B1}{180}
    \wirelesssymbol{B2}{0}

    \node[above=0.675cm of pc.north, xshift=-0.25cm] (C1) {};
    \node[above=0.75cm of C1] (C2) {};
    \wirelesssymbol{C1}{90}
    \wirelesssymbol{C2}{270}

    \node[above=0.675cm of pc.north, xshift=0.25cm] (D1) {};
    \node[above=0.75cm of D1] (D2) {};
    \wirelesssymbol{D1}{90}
    \wirelesssymbol{D2}{270}

    \draw[arrow] ($(spot.south) + (0.25cm,0)$) -- node[right,align=left]{Pose\\estimate} ($(spot.south |- A2) + (0.25cm,0)$) -- ($(A2)$);
    \draw[arrow] ($(A1)$) -- ($(pc.west) + (0,0.25cm)$);
    
    \draw[arrow] ($(B2)$) -- ($(spot.south |- B2) + (-0.25cm,0)$) -- node[left,align=right]{Pose\\reference} ($(spot.south) + (-0.25cm,0)$);
    \draw[arrow] ($(pc.west) + (0,-0.25cm)$) -- ($(B1)$);

    \draw[doublearrow] ($(C1)$) -- node[left]{RGB-D} ($(pc.north) + (-0.25cm,0)$);
    \draw[doublearrow] ($(rpi.south) + (-0.25cm,0)$) -- ($(C2)$);

    \draw[arrow] ($(pc.north) + (0.25cm,0)$) -- node[right,align=left]{Tool speed\\reference} ($(D1)$);
    \draw[arrow] ($(D2)$) -- ($(rpi.south) + (0.25cm,0)$);

    \begin{pgfonlayer}{background}
    \node[componentbox, fit=(tool)(camera)(battery)(rpi)(C2), label={[anchor=south east, align=right]south east:\textbf{Onboard}\\\textbf{components}}] (onboard) {};
    \end{pgfonlayer}

    \node[anchor=south east, align=left, fill=lightgray!50,
    inner sep=4pt] at (current bounding box.south east) {%
    \textbf{Legend}\\
    \tikz{\draw[arrow] (0,0)--(0.5,0);} Data\\
    \tikz{\draw[arrow,dashed] (0,0)--(0.5,0);} Power};
    
    \end{tikzpicture}
    }
    
    \caption{The architecture with remote processing.}
    \label{fig:integration-rpi}

    \end{subfigure}

    \vspace{4mm}

    \begin{subfigure}{\columnwidth}
    \centering

    \resizebox{\linewidth}{!}{%
    \begin{tikzpicture}[
    box/.style={draw, thick, fill=white, minimum width=2.0cm, minimum height=1.0cm, align=center},
    smallbox/.style={draw, thick, minimum width=2.4cm, minimum height=1.2cm, align=center},
    arrow/.style={thick, -{Stealth[length=2mm,width=2mm]}},
    doublearrow/.style={thick, {Stealth[length=2mm,width=2mm]}-{Stealth[length=2mm,width=2mm]}},
    componentbox/.style={draw=none, fill=AccentBlue!70, inner sep=0.5cm},
    wireless/.style={draw, thick, line cap=round}
    ]

    \node[box] (spot) {Spot};

    \node[box, right=3.75cm of spot] (jetson) {Jetson\\AGX Orin};

    \node[box, above left=1.0cm and -0.5cm of jetson] (tool) {Weeding\\Tool};
    \node[box, above right=1.0cm and -0.5cm of jetson] (camera) {Camera};
    
    \draw[arrow,dashed] (spot.north) -- ($(spot.north |- tool.west) + (0.0, -0.7)$) -- ($(spot.north |- tool.west) + (1.2, -0.7)$) -- ($(spot.north |- tool.west) + (1.2, 0.0)$) -- node[above]{} (tool.west);
    \draw[arrow,dashed] ($(spot.east)$) -- ($(jetson.west)$);
    \draw[arrow,dashed] (jetson.east) -- node[right]{} ($(camera.south |- jetson.east) + (0.5cm,0)$) -- ($(camera.south) + (0.5cm,0)$);
    
    \draw[arrow] ($(tool.south |- jetson.north) + (0.75cm,0)$) -- node[left, align=right]{Tool speed\\reference} ($(tool.south) + (0.75cm,0)$);
    
    \draw[arrow] ($(camera.south) + (-0.75cm,0)$) -- node[left]{RGB-D} ($(camera.south |- jetson.north) + (-0.75cm,0)$);

    \draw[arrow] ($(spot.east) + (0.0cm,0.25cm)$) -- node[pos=0.25,above,align=left]{Pose\\estimate} ($(jetson.west) + (0.0cm,0.25cm)$);
    
    \draw[arrow] ($(jetson.west) + (0.0cm,-0.25cm)$) -- node[pos=0.75,below,align=left]{Pose\\reference}($(spot.east) + (0.0cm,-0.25cm)$);
    
    \begin{pgfonlayer}{background}
    \node[componentbox, fit=(tool)(camera)(jetson), label={[anchor=south east, align=right]south east:\textbf{Onboard}\\\textbf{components}}] (onboard) {};
    \end{pgfonlayer}

    \node[anchor=north west, align=left, fill=lightgray!50,
    inner sep=4pt] at (current bounding box.north west) {%
    \textbf{Legend}\\
    \tikz{\draw[arrow] (0,0)--(0.5,0);} Data\\
    \tikz{\draw[arrow,dashed] (0,0)--(0.5,0);} Power};
    
    \end{tikzpicture}
    }
    
    \caption{The architecture with onboard processing.}
    \label{fig:integration-jetson}
    \end{subfigure}

    \caption{The hardware architectures used. Two sided arrows represent data received after a request.}
    \label{fig:integration}
    
\end{figure}

\section{Control architecture}
\label{sec:control-architecture}
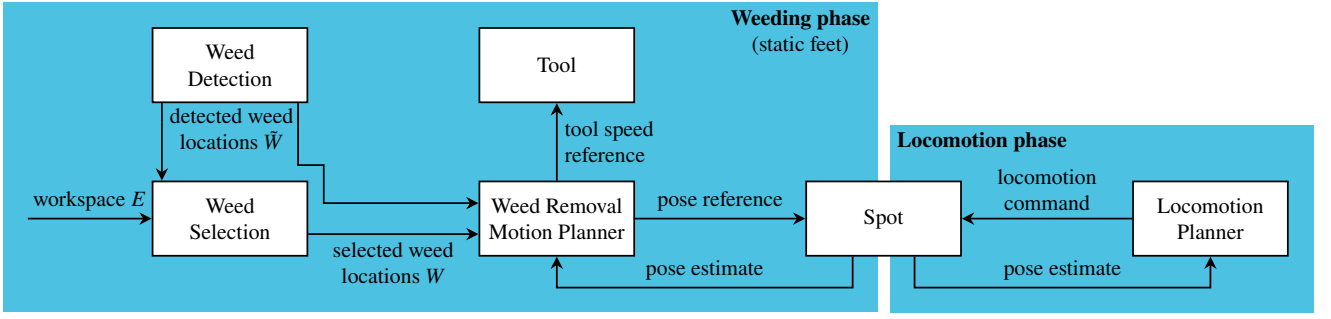
\begin{figure*}[!btp]
    \centering

    \resizebox{\linewidth}{!}{%
    \begin{tikzpicture}[
    box/.style={draw, thick, fill=white, minimum width=2.5cm, minimum height=1.2cm, align=center},
    smallbox/.style={draw, thick, minimum width=2.5cm, minimum height=1.2cm, align=center},
    arrow/.style={thick, -{Stealth[length=2mm,width=2mm]}},
    doublearrow/.style={thick, {Stealth[length=2mm,width=2mm]}-{Stealth[length=2mm,width=2mm]}},
    componentbox/.style={draw=none, fill=AccentBlue!70, inner sep=0.4cm},
    wireless/.style={draw, thick, line cap=round}
    ]

    \node[box] (selection) {Weed\\Selection};
    \node[box, above=1.25cm of selection] (detection) {Weed\\Detection};
    \node[box, right=2.75cm of selection] (removal) {Weed Removal\\Motion Planner};
    \node[box, right=2.75cm of removal] (spot) {Spot};
    \node[box, right=2.75cm of spot] (locomotion) {Locomotion\\Planner};
    \node[box, above=1.25cm of removal] (tool) {Tool};

    \coordinate[below=0.5cm of spot, xshift=-0.5cm] (spot1);
    \coordinate[below=0.5cm of spot, xshift=0.5cm] (spot2);
    \coordinate[above=0.5cm of spot, xshift=0.5cm] (spot3);
    \coordinate[left=2.0cm of selection] (workspace);

    \draw[arrow] ($(detection.south)+(-1.1cm,0)$) -- node[right, pos=0.33, align=center]{detected weed\\locations $\tilde{W}$} ($(selection.north)+(-1.1cm,0)$);
    \draw[arrow] ($(selection.east) + (0,-0.25cm)$) -- node[below,align=center]{selected weed\\locations $W$} ($(removal.west) + (0,-0.25cm)$);
    \draw[arrow] ($(detection.south)+(1.1cm,0)$) -- ($(selection.north)+(1.1cm,0.25cm)$) -- ($(selection.east |- selection.north) + (0.25cm,0.25cm)$) -- ($(selection.east) + (0.25cm,0.25cm)$) -- ($(removal.west) + (0,0.25cm)$);
    \draw[arrow] (workspace) -- node[above,align=left]{workspace $E$} (selection.west);
    \draw[arrow] (removal.east) -- node[above,align=left]{pose reference} (spot.west);
    \draw[arrow] (locomotion.west) -- node[above,align=center]{locomotion\\command} (spot.east);
    \draw[arrow] (removal.north) -- node[right,align=left]{tool speed\\reference} (tool.south);
    \draw[arrow] ($(spot.south) + (-0.5cm,0)$) -- ($(spot.south) + (-0.5cm,-0.5cm)$) -- node[above,align=left]{pose estimate} ($(removal.south) + (0,-0.5cm)$) -- (removal.south);
    \draw[arrow] ($(spot.south) + (0.5cm,0)$) -- ($(spot.south) + (0.5cm,-0.5cm)$) -- node[above,align=left]{pose estimate} ($(locomotion.south) + (0,-0.5cm)$) -- (locomotion.south);

    \begin{pgfonlayer}{background}
    \node[componentbox, fit=(workspace)(tool)(detection)(selection)(removal)(tool)(spot1), label={[anchor=north east, align=center]north east:\textbf{Weeding phase}\\(static feet)}] {};
    \node[componentbox, fit=(locomotion)(spot2)(spot3), label={[anchor=north west, align=center]north west:\textbf{Locomotion phase}}] {};
    \end{pgfonlayer}

    \end{tikzpicture}
    }

    \caption{Schematic overview of the architecture used to control the setup, containing two phases: the weeding phase and the locomotion phase.}
    \label{fig:control-architecture}
\end{figure*}

The proposed control architecture is shown in \autoref{fig:control-architecture}. It alternates between the locomotion phase and the weed removal (with static feet) phase. The main reason for splitting the two phases is to prevent disturbances caused by locomotion to negatively influence the performance of the weed removal. Weeding can still be achieved by moving the body using the degrees of freedom (DOFs) in the legs while the feet remain at static positions on the ground. The locomotion phase is meant to bring new weeds into the robot's workspace. During the weed removal phase, the system first receives the weed locations from the camera and verifies whether they are within the workspace $E$, after which it determines the order of weed removal. The weeds are then removed one by one by the weed removal procedure. In the subsequent sections, several key aspects of the implementation are discussed in more detail.

\subsection{Workspace}
We define the workspace as the set of $(x, y, z)$ positions of the end-effector (EF) for which there exist feasible joint angles to reach this position without repositioning the feet.

\subsubsection{Degrees of freedom}
\label{sec:dof}
Every leg of Spot has three actuated DOFs: two in its hips and one in its knees. This means that even when the feet positions are stationary, Spot's body is capable of movements in $\mathrm{SE}(3)$ with limited range, allowing translation along the $x$, $y$, and $z$ axes, and rotations about these axes by angles $\phi$, $\theta$, and $\psi$, respectively (ordered as $z$, $y$, $x$ Euler angles), with respect to Spot's body frame as defined in \autoref{fig:spot-body-frame}. To simplify the inverse kinematics and achieve a unique closed-form solution (within the workspace), only three of the six available DOFs are utilized to move the EF to a specified location in $\mathbb{R}^3$.

Translation along the $z$ axis is excluded, as this motion alone does not enable ground contact for the EF; the joint range is used more effectively for pitching. Translation along the $y$ axis is also excluded due to its limited range before the robot falls over; a larger lateral movement of the EF can be achieved by yawing. Finally, roll (rotation about the $x$ axis) is disregarded, since rolling the body increases the vertical distance between the ground and the EF, hindering its ability to touch the ground. Consequently, the DOFs retained to achieve the desired position of the EF are: translation along the $x$ axis; rotation about the $y$ axis (pitch $\theta$); rotation about the $z$ axis (yaw $\psi$).

\subsubsection{Forward and inverse kinematics}
\label{sec:kinematics}
Using the selected DOFs as inputs, the forward kinematics (FK) are used to find the position \mbox{$p_{\text{EF}} = \begin{bmatrix}x_{\text{EF}}, y_{\text{EF}}, z_{\text{EF}}\end{bmatrix}^\top$} of the EF based on $x$, $\theta$ and $\psi$ and the dimensions of the tool ($x_{\text{arm}}$, $y_{\text{arm}}$ and $z_{\text{arm}}$, the dimensions indicating the EF location in the body fixed frame) with respect to the center of the body. The transformation matrix from the body fixed frame to the world fixed frame representing this movement is given by:
\begin{equation}
    \begin{bmatrix}
        x_{\text{EF}} \\
        y_{\text{EF}} \\
        z_{\text{EF}} \\
        1
    \end{bmatrix}
    =
    \begin{bmatrix}
        c_\psi c_\theta & - s_\psi & c_\psi s_\theta & x \\
        s_\psi c_\theta &   c_\psi & s_\psi s_\theta & 0 \\
        - s_\theta      & 0        & c_\theta        & z_{\text{nom}} \\
        0 & 0 & 0 & 1
    \end{bmatrix}
    \begin{bmatrix}
        x_{\text{arm}} \\
        y_{\text{arm}} \\
        z_{\text{arm}} \\
         1
    \end{bmatrix}
\end{equation}
where $c_{\alpha}$ and $s_{\alpha}$ represent $\cos({\alpha})$ and $\sin({\alpha})$, respectively, with \mbox{$\alpha \in \{\psi,\theta\}$}, and $z_{\text{nom}} = \SI{0.5}{\meter}$ is the nominal height of the body with respect to its feet on the ground when providing a body pose command of 0-height. Given a desired EF position, the objective is to find the input values $x$, $\psi$ and $\theta$ that position the EF at the target location. By solving the inverse kinematics (IK) and taking \mbox{$y_{\text{arm}} = 0$} (due to the tool's center alignment), the following equations are obtained:
\begin{align}
    \theta &= \arccos\left(\frac{z_{\text{EF}}-z_{\text{nom}}}{\sqrt{x_{\text{arm}}^2+z_{\text{arm}}^2}}\right)-\arctan\!2\!\left(x_{\text{arm}},z_{\text{arm}}\right), \label{eq:theta-ik}\\
    \psi &= \arcsin\left(\frac{y_{\text{EF}}}{c_\theta x_{\text{arm}} + s_\theta z_{\text{arm}}} \right), \label{eq:psi-ik} \\
    x &= x_{\text{EF}} - c_\psi c_\theta x_{\text{arm}} - c_\psi s_\theta z_{\text{arm}}. \label{eq:x-ik}
\end{align}

\subsubsection{Kinematically and experimentally feasible workspaces}
\begin{figure}
  \centering
  \begin{subfigure}{\columnwidth}
    \centering
    \includegraphics[width=\linewidth]{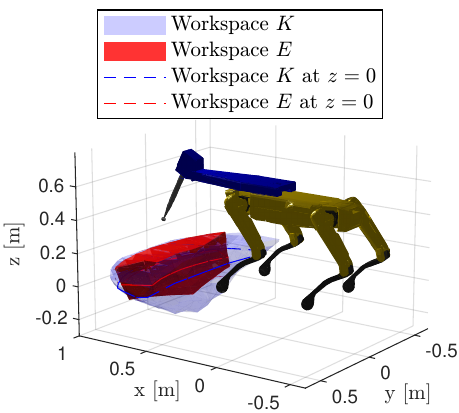}
    \caption{3D workspace of Spot with the custom weed removal tool. Although the workspace extends further along the positive $z$-axis, this extension is not depicted for clarity.}
    \label{fig:3D-workspace}
  \end{subfigure}
  
  \vspace{4mm}
  
  \begin{subfigure}{\columnwidth}
    \centering
    \includegraphics[width=\linewidth]{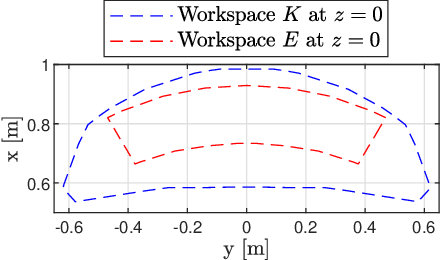}
    \caption{Cross-cut of the workspaces at $z = 0$.}
    \label{fig:2D-workspace}
  \end{subfigure}
  \caption{Different representations of the workspaces.}
  \label{fig:workspace}
\end{figure}

The equations presented are valid only when the desired pose of the EF lies within Spot's workspace. Let $K\subset\mathbb{R}^3$ denote the \textbf{kinematically feasible workspace}, defined as the set of EF positions $(x,y,z)\in\mathbb{R}^3$ for which there exist joint angles $\theta\in\Theta$, where $\Theta$ contains the kinematically feasible joint angle ranges for each joint, that satisfy the kinematic constraints of the legs, allowing the EF to reach the desired position without the need to reposition the feet. This workspace $K$ can be determined analytically using the full body kinematics reported in \autoref{app:inverse-kinematics}, and is illustrated in \autoref{fig:3D-workspace}.

However, in Spot, the joint limits are not solely dictated by the kinematic constraints of the legs, but also influenced by an internal controller, of which the specifics are undisclosed. As a result, the actual workspace, denoted by \mbox{$E~\subset~\mathbb{R}^3$}, can only be accurately characterized through empirical validation. The \textbf{experimentally feasible workspace} $E$ is thus defined as the set of EF positions \mbox{$(x,y,z)\in\mathbb{R}^3$} for which there exist internally feasible joint angles \mbox{$\theta^* \in \Theta^*$}, where $\Theta^*\subseteq\Theta$, which allow the EF to reach the position without requiring foot repositioning. Workspace $E$, in the case of Spot standing on flat ground, is found experimentally and illustrated in \autoref{fig:3D-workspace}. Note that \mbox{$E~\subseteq~K$}.

To further visualize the difference between the kinematically feasible workspace $K$ and the experimentally feasible workspace $E$, cross-sections of both workspaces at $z=0$ (ground level) are presented in \autoref{fig:2D-workspace}. The surface area of $K$ at $z=0$ equals \SI{0.402}{\meter\squared}, whereas the surface area of $E$ at $z=0$ equals \SI{0.165}{\meter\squared}. Only 41\% of the area kinematically feasible is still feasible using Spot's internal controller when the ground is flat.

The workspaces analyzed so far represent $K$ and $E$ when Spot is positioned on a flat, horizontal surface. In scenarios where the terrain is uneven and/or tilted, conditions frequently encountered in agricultural fields, the workspaces $K$ and $E$ are further reduced to $\hat{K}$ and $\hat{E}$, respectively, due to the increased joint range required to maintain Spot's stability. A relevant workspace $\hat{E}$ cannot be found for every scenario, as this can only be found empirically. For some scenarios, the workspace $\hat{E}$ is found and visualized in \autoref{app:workspace}. In \autoref{tab:reduced-workspace}, the area of the reduced workspaces $\hat{K}$ and $\hat{E}$ at $z = 0$ is reported for a variety of scenarios. In all cases, $\hat{E}$ is smaller than $E$.

In the control pipeline, workspace $E$ is used to check if weed removal is feasible, even though the actual workspace might be smaller in some cases, during which weed removal will fail. The fact that $E$ is much smaller than $K$ and that $\hat{E}$ cannot be found analytically is a major limitation of the Spot system and can only be resolved by switching to a different system that reports the exact specifications of the internal controller or by switching to a system that allows joint control, allowing us to determine the limits ourselves.

\begin{table}[tb]
    \caption{The area in \si{\meter\squared} of the reduced workspaces at $z=0$ for different scenarios. The columns FL (Front Left), FR (Front Right), HL (Hind Left), HR (Hind Right) contain how much each foot is raised with respect to the ground in \si{\meter}.}
    \label{tab:reduced-workspace}
    \begin{tabular*}{\tblwidth}{@{} LRRRRRR@{} }
    \toprule
        Case & FL & FR & HL & HR & $\hat{K}$ at $z = 0$ & $\hat{E}$ at $z = 0$ \\
        \midrule
        1 & 0.05 & 0.00 & 0.00 & 0.00 & 0.420 & 0.111 \\
        2 & 0.05 & 0.10 & 0.00 & 0.00 & 0.455 & 0.012 \\
        3 & 0.05 & 0.00 & 0.10 & 0.00 & 0.459 & 0.058 \\
        4 & 0.05 & 0.00 & 0.00 & 0.10 & 0.557 & 0.113 \\
        \bottomrule
    \end{tabular*}
\end{table}

\subsection{Weed detection}
\label{sec:weed-detection}
The Intel RealSense L515 camera, as introduced in \autoref{sec:camera-setup}, is used to detect weeds and extract their location in 3D. The depth and RGB images are aligned, and by utilizing the camera intrinsics to correct radial distortion, the 3D position of pixels within the image can be determined based on a given pixel location and its corresponding depth value.

\begin{figure}
  \centering
  \begin{subfigure}{\columnwidth}
    \centering
    \includegraphics[width=\linewidth]{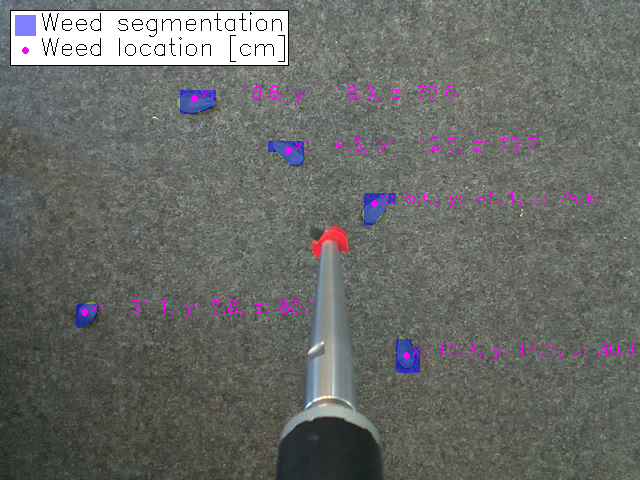}
    \caption{YOLOv8 segmentation results and corresponding weed locations in the camera frame for artificial weeds.}
    \label{fig:weed-fake}
  \end{subfigure}
  
  \vspace{4mm}
  
  \begin{subfigure}{\columnwidth}
    \centering
    \includegraphics[width=\linewidth]{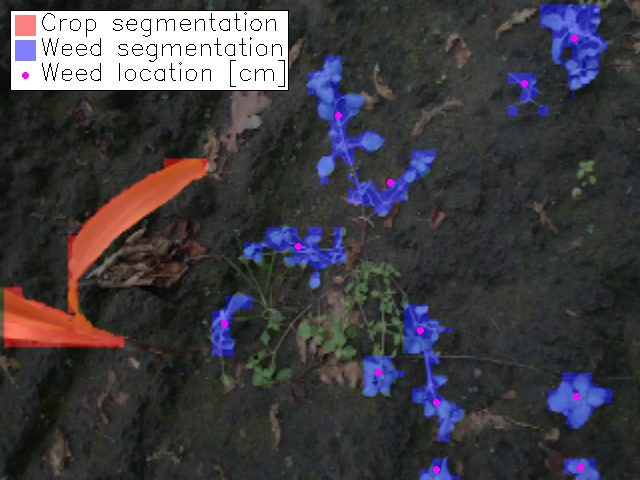}
    \caption{YOLOv8 segmentation results and corresponding weed locations in the camera frame for real weeds.}
    \label{fig:weed-real}
  \end{subfigure}
  \caption{YOLOv8 segmentation results.}
  \label{fig:weed-detection}
\end{figure}

Weed detection is achieved using a trained YOLOv8 segmentation network \citep{glenn2023yolov8}, which is trained on RGB data. Two distinct networks are utilized: one trained on artificial indoor weed data, the other on real outdoor data collected from a maize field, with the aim of detecting and classifying both weeds and maize plants. The target point for weed removal $w$, containing $x$, $y$, and $z$ in the camera frame, is identified as the centroid of the segmented weed mask. All weed locations are stored in the unordered weed vector $\tilde{W}$. Example detections and their corresponding removal locations for both artificial and real weeds can be seen in \autoref{fig:weed-fake} and \autoref{fig:weed-real}, respectively.

To prevent the weed removal tool's shaft from being misidentified as a weed, it is masked in the image. Although the exact location of the camera casing relative to the tool is known, the precise location of the lens's center point is not. The lens is assumed to be perpendicular to the camera casing. To accurately determine the center point of the lens, a reference point on the shaft of the removal tool is selected. Given that the coordinates of this reference point relative to the center point of the camera casing are known, the camera lens can be calibrated to align with the camera casing.

\subsection{Weed removal motion planner}
The removal planner determines which weeds in $\tilde{W}$ detected by the camera to remove and in what order. The removal planner receives the unordered weed vector $\tilde{W}$ from the weed detection module on request. Only weeds that are within the workspace $E$ are stored:
\begin{equation}
    \hat{W} = \{w \in \tilde{W} \mid w \in E\}.
\end{equation}
For all weeds \mbox{$w \in \hat{W}$}, the desired removal pose $p$, containing a desired $x$, $\theta$, and $\psi$, is calculated using Equations \eqref{eq:theta-ik}--\eqref{eq:x-ik}. The weed locations \mbox{$w \in \hat{W}$} are sorted in increasing order of $\psi$, resulting in the ordered weed vector $W$. This determines the order of removal, corresponding to a removal yawing from right to left. This type of removal is chosen to prevent the gradual outward creep of the front feet that results from the constant yawing from left to right and back.

Since the camera's field of view does not cover the entire workspace $E$, the static feet removal step is performed twice. In one iteration the scan is made with \mbox{$\psi=\SI[parse-numbers=false]{-\ \nicefrac{\pi}{18}}{\radian}$}, and in the other iteration \mbox{$\theta=\SI[parse-numbers=false]{+\ \nicefrac{\pi}{18}}{\radian}$} is used, ensuring that the entire workspace is covered. Some overlap exists between the camera frames in the two iterations, but weeds visible in the first frame are (attempted to be) removed before the second reading takes place.

\subsection{Weed removal}
\label{sec:removal}
The weed removal step receives weed locations \mbox{$w \in W$} from the removal planner one by one and is responsible for removing them. Spot's internal controller is designed to receive a target pose \mbox{$p \in \mathrm{SE}(3)$}, containing translations ($x$,$y$,$z$) and rotations ($\phi$,$\theta$,$\psi$) for the body frame (see \autoref{sec:dof}), and track it, with the objective of reaching and then stationary maintaining that pose. As a result, computing and following a complete trajectory is unnecessary and even undesirable, as it would cause Spot to briefly halt at each intermediate pose. Instead, the desired final pose $p$ that has an end-effector location corresponding to $w$ is directly communicated to Spot to optimize its motion efficiency.

Inevitably, there are disturbances and inaccuracies that cause the location of the end-effector at $p$ to misalign with $w$. This requires an update of the pose before removing the weed, using a feedback step. The main sources of disturbance and inaccuracy we identified that cause an undesirable difference between the end-effector and $w$ are: (1) due to imperfect camera readings, the extracted weed location $w$ will differ from the actual weed location $w^*$, which is unknown, but the closer the camera moves to the weed, the better $w$ estimates $w^*$; (2) the pose $p$ might not be tracked perfectly by Spot's internal controller; and (3) once the robot moves closer to the weed, it pitches forward and more weight is put onto the front feet, which can cause the feet to sink deeper into the soft ground, resulting in a change of $w$ in the body frame.

To account for these disturbances and inaccuracies, we use an adaptive target updating controller that uses visual feedback from the camera to adapt the target pose $p$. This controller receives the unordered weed vector $\tilde{W}$ from the camera and finds the weed $w_{\text{new}}$ that is closest to $w$ in the initial body frame. If this distance is larger than \SI{10}{\centi\meter}, it is assumed that the same weed has not been detected again and the old weed location $w$ in the initial body frame will be used to update the pose $p$.

The resulting correction step first computes the EF position based on the current target pose using FK, adds the remaining offset between the weed and the EF, and then uses IK to obtain a new target pose:
\begin{equation}
    p_{\text{new}} = \mathrm{IK}(\mathrm{FK}(p_{\text{current}}) + w_{\text{new}}-p_{\mathrm{EF}}),
\end{equation}
where $\mathrm{IK}$ and $\mathrm{FK}$ denote inverse kinematics and forward kinematics, respectively, and $p_{\mathrm{EF}}$ represents the end-effector position (see \autoref{sec:kinematics}). All positions are expressed in the initial body frame.

In this controller, $p_{\text{current}}$ and $\tilde{W}$ must be measured at the same time instance, or the robot should not move between these points in time, as a deviation in time will put the measurement of $\tilde{W}$ in the wrong position when the robot moved. With onboard processing (\autoref{fig:integration-jetson}), this is no problem, but when performing the calculations on a remote system (as in \autoref{fig:integration-rpi}), we cannot guarantee that these measurements are time-synchronized. In addition, receiving the camera data and running the neural network might take longer than the movement. Therefore, we close the loop only once and receive a camera reading when the robot is stationary with the end-effector just above and behind or at the side of the expected weed location, preventing potential obstruction of the camera's view of the weed by the tool, to guarantee that the pose and camera reading estimates are time-synchronized. In addition, camera images can suffer from motion blur, making detection when stationary more beneficial.

\subsection{Locomotion planner}
The locomotion planner is responsible for positioning Spot in a pose \mbox{$r \in \mathrm{SE}(2)$}, containing the desired values $x$, $y$, and $\psi$ to which Spot should move with respect to the current body frame. Developing a sophisticated planner is outside the scope of this research. Instead, a simple locomotion planner is made that commands Spot to walk forward in a straight line using the velocity command functionality in the API, covering approximately \SI{20}{\centi\meter} at each step, which corresponds to the depth of the workspace. In this way, an approximately \SI{80}{\centi\meter} wide strip is covered, corresponding to the approximate width of the workspace $E$.

\section{Results}
\label{sec:results}
In this section, the results of the introduced control architecture in combination with the setup and integration are discussed in three environments: simulation, indoor with artificial plants, and outdoor with real plants. For practical and season-dependent reasons, all the results in this section were obtained with the remote processing architecture shown in \autoref{fig:integration-rpi}.

\subsection{Simulation}
\begin{figure*}
    \centering
    \includegraphics[width=\linewidth]{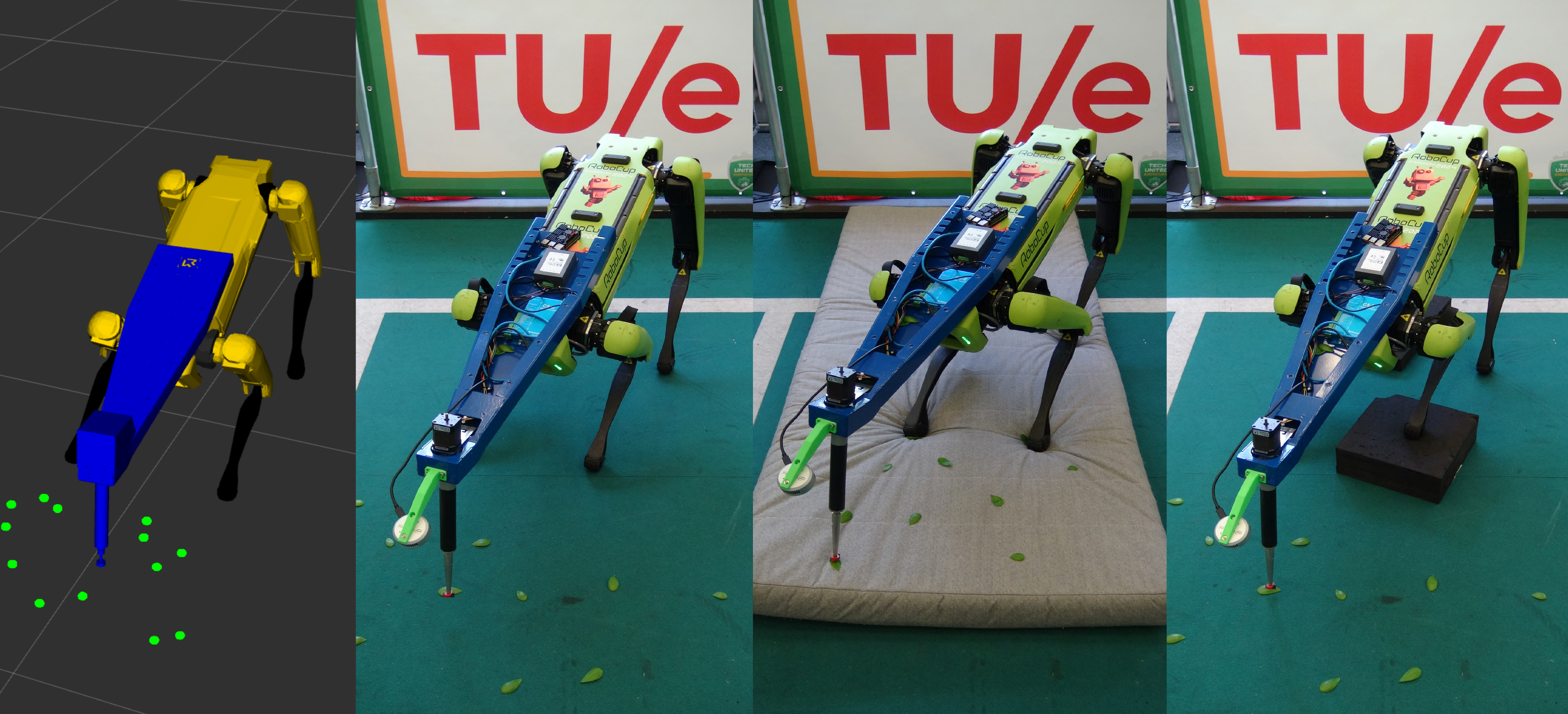}
    \caption{Testing the system in simulation and indoors in different difficulty levels, from left to right: simulation, indoors flat ground, soft ground, and uneven ground.}
    \label{fig:simulated-results}
\end{figure*}

A dedicated simulation environment has been developed to facilitate the testing of various software versions. This environment includes a Spot robot equipped with the custom weed removal tool and is based on the full-body kinematics. In addition, a simulated camera is capable of detecting randomly initialized weeds. The environment, shown on the left in \autoref{fig:simulated-results}, provides a complete platform for extensive testing of the removal path planner and the adaptive target updating controller. Additionally, the environment allows for the evaluation of the impact of minor misalignments in the tool attachment to Spot, particularly relevant in the unconstrained $x$-direction, where deviations from the correct position can occur. Given that the camera is mounted on the tool, a misalignment of \SI{3}{\centi\meter} in the tool's position with respect to Spot's body results in an end-effector error of less than \SI{3}{\milli\meter}. This level of accuracy indicates that the system can tolerate slight inaccuracies during the tool mounting process.

\subsection{Indoor}
The indoor experiments are used to evaluate the system's ability to accurately track desired inputs and to assess the performance of the entire pipeline, including camera-based artificial weed detection. These tests were conducted in various scenarios, as illustrated in \autoref{fig:simulated-results}. Initially, the system is tested on a flat ground as the simplest use case. Subsequent tests are conducted on a mattress to simulate the soft ground conditions typically found in real agricultural fields. The system's robustness was further challenged by testing it with some of Spot's feet elevated, to assess the controller's ability to handle uneven terrain. In all scenarios, the system demonstrated effective weed removal capabilities.

Extensive indoor testing on the mattress is conducted to thoroughly analyze the system's performance. The success rate of the system, defined as the percentage of detected weeds that were successfully removed, is given in \autoref{tab:success-rate}. The times required to execute each component of the code are presented in \autoref{tab:times}.

The time required to clear a specific area of weeds is directly related to the density of weeds in the field. For instance, with an estimated density of 5 weeds per workspace, which is considered realistic \citep{Wallinga1997}, one complete cycle of the entire pipeline takes \SI{47}{\second}, covering an area of \SI{0.165}{\meter\squared}. Consequently, clearing an entire hectare (\SI[parse-numbers=false]{10,000}{\meter\squared}) would require approximately 790 hours using this system. To match the efficiency of comparable systems such as Odd.Bot and Trabotyx, which can clear a hectare in 16 to 18 hours, respectively, approximately 46 Spot robots would be necessary.
The field could also be continuously monitored and fully cleared over the course of one week, which corresponds to the time it takes for weeds to progress from emergence to a problematic stage. Under these conditions, 5 Spot robots would be required, assuming that the robots are operational about 95\% of the time, with the remaining 5\% allocated to battery replacement and maintenance.

\begin{table}[ht]
    \caption{Removal success rate for detected weeds. Since these results come from indoor tests, the artificial weeds are not physically removed by the milling bit, but they are successfully touched at the desired location. Italic values represent individual components that sum to their respective total value.}
    \label{tab:success-rate}
    \begin{tabular*}{\tblwidth}{@{} LRR@{} }
        \toprule
         & Number & Percentage \\
        \midrule
        Removed                 & 72 & 82\%\\
        Not Removed (Total)     & 16 & 18\%\\
        \hspace{5mm}\textit{Poor tracking}        & \textit{12} & \textit{14\%}\\
        \hspace{5mm}\textit{No second detection}  & \textit{0} & \textit{0\%}\\
        \hspace{5mm}\textit{Outside $E$}          & \textit{4} & \textit{4\%}\\
        \bottomrule
    \end{tabular*}
\end{table}

\begin{table}[ht]
    \caption{Required time for a single iteration of each part of the pipeline, and how many iterations are required for each part per cycle, where $n$ represents the number of weeds in the workspace. Italic values represent individual components that sum to their respective total value.}
    \label{tab:times}
    \begin{tabular*}{\tblwidth}{@{} LRR@{} }
        \toprule
         & Time [s] & Iterations\\
        \midrule
        Removal planner                  & 2.384 & 2\\
        Weed removal per weed (Total)    & 7.789 & $n$\\
        \hspace{5mm} \textit{Movement}   & \textit{4.795} & $n$ \\
        \hspace{5mm} \textit{Detection}  & \textit{1.994} & $n$\\
        \hspace{5mm} \textit{Removal}    & \textit{1.000} & $n$ \\
        Locomotion planner               & 3.151 & 1\\
        \bottomrule
    \end{tabular*}
\end{table}

\subsection{Outdoor}
\begin{figure*}[!b]
    \centering
    \includegraphics[width=\linewidth]{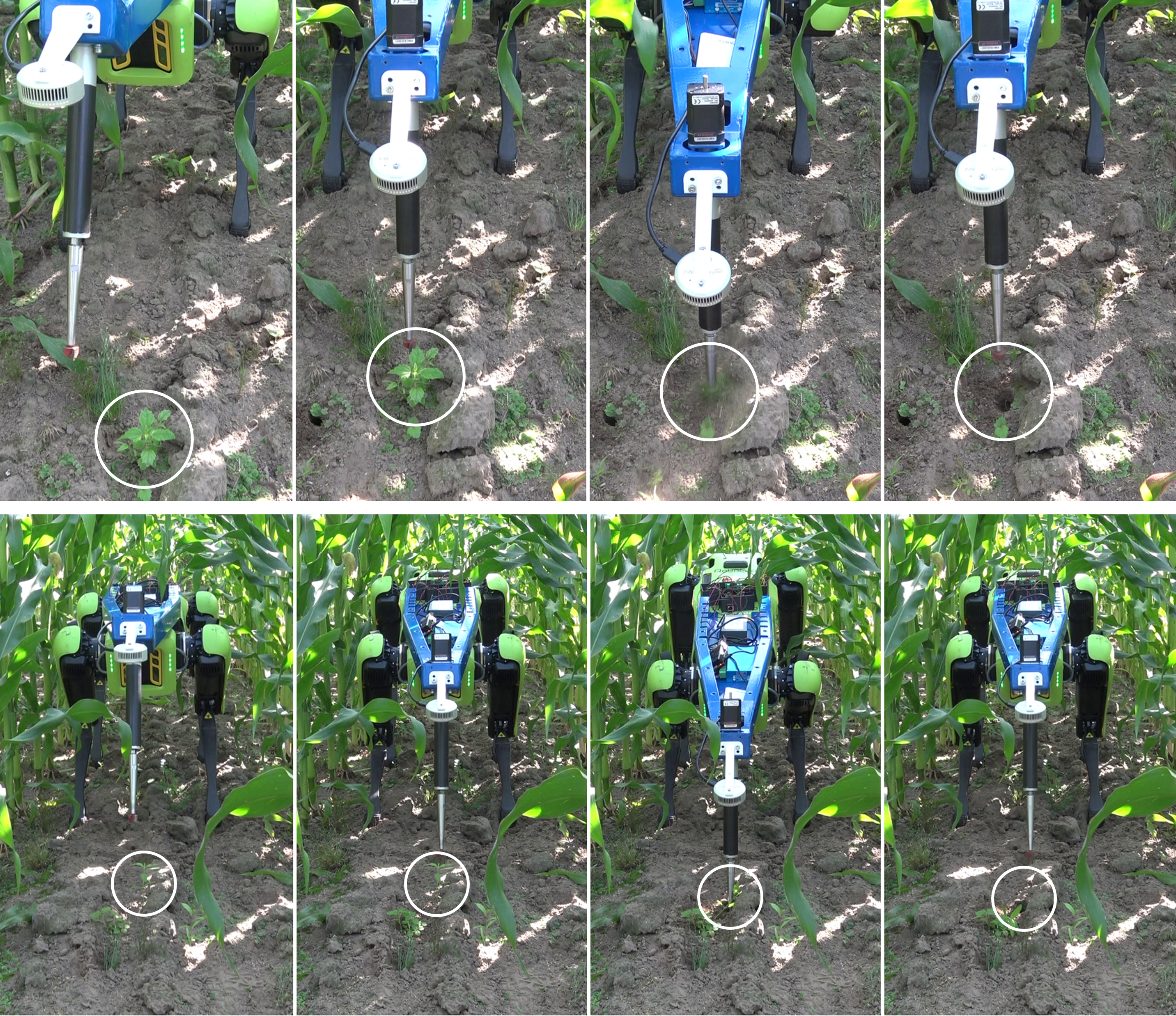}
    \caption{Two scenarios of successful weed removal attempts outdoors. From left to right, the images represent the removal path planner step, the weed removal step during feedback, the final weed removal step, and the resulting removal of the plant.}
    \label{fig:test-outdoor}
\end{figure*}

To evaluate the technical feasibility of the weed removal tool and assess the performance of the entire pipeline in real-world conditions, the system was subjected to outdoor testing. These tests were conducted in a maize field with fully grown crops, as depicted in \autoref{fig:test-outdoor}.
In this scenario, the weeds were larger than those typically encountered during the pre-canopy closure period. During early crop growth, weeds are typically small and further development can be prevented by removing them. After canopy closure, weeds are harder to reach and detect, both for humans and conventional machinery within the dense, fully grown canopy, but they are also less problematic, even if they grow larger. However, removing the plants can still be beneficial: if done before they set seed, it prevents them from adding to the weed seed bank, and for perennial weeds, which can also spread vegetatively through their roots, removal additionally helps limit this further growth and spread. Both effects can reduce weed pressure in the next growing season. As our quadruped platform can still access this environment, testing it under these more challenging conditions provides a valuable demonstration of its capability, complementary to its primary intended use during the pre-canopy closure period.

The outdoor tests presented additional challenges, particularly due to the varying lighting conditions compared to those present during the initial data collection phase. As a result, the weed detection algorithm, which relies on a YOLO detection network, did not function as effectively under these new conditions as it did under lab conditions. As our main focus in this research was to test the proposed weed mechanism rather than perfect detection (see \autoref{sec:requirements}), real-time manual detection was employed to identify weeds during the testing process, so that the results are judged independently of detection mistakes.

The results of the outdoor tests were not as successful as their indoor counterpart, with the system achieving complete successful weed removal in approximately 25\% of the scenarios. Notably, the tracking error observed during outdoor testing was similar to that recorded in the indoor environment, indicating that tracking accuracy was not the primary cause of the reduced performance.

The primary issue encountered during outdoor testing was related to accurately determining the correct removal locations. Although the camera was capable of detecting weeds, the system struggled to precisely identify and target the removal points. The weeds in the test field were relatively tall, reaching heights of up to \SI{10}{\centi\meter}, and often did not grow straight, making it difficult to accurately locate the weed's roots. The size of the weeds encountered during testing exceeded the parameters for our intended application, which is optimized for removal during the pre-canopy closure period with smaller weeds. Additionally, the system's tool does not descend in a straight vertical line; instead, it utilizes a pitching movement during descent. This motion results in overshooting the target above the ground, and only reaching the roots possibly somewhere underground. Still, the successful weed removal under real conditions in the cases when the removal points were correctly detected reveal that the system can provide an effective solution for weed removal.

\section{Conclusion and discussion}
\label{sec:conclusions}
This work explores the use of a mobile robotic quadruped platform equipped with a tool for mechanical weed removal in agricultural fields, presenting a viable alternative to traditional herbicide-based weed management methods. Apart from removing the need for herbicides, and thus providing a more sustainable method of weed removal, this method offers several other advantages, such as reduced soil compaction, easier scalability, safer autonomy, and the possibility to integrate this approach with intercropping. 

To this end, we have equipped a Boston Dynamics Spot robot with a custom-made 0-DOF weed removal tool, featuring a milling bit on its end used for destroying weeds that are detected by a RGB-D camera mounted above the tool. The milling bit is positioned at the desired location using the degrees of freedom in Spot's body while keeping its feet stationary. This allows for inter-row as well as intra-row weed removal for a larger amount of field conditions and field types. A hardware and software structure were created, to explore requirements and challenges. Additionally, the used system was analyzed to identify limitations on e.g. the workspace and accuracy.

The complete system has been analyzed in simulation, indoors, and outdoors, in order to assess its practical feasibility. Indoor tests show that a removal accuracy higher than 80\% can be achieved when the right removal location is found. Outdoor tests confirm the mechanical feasibility of the tool, but also highlight some important steps that need to be taken to make such a solution fully operational. Namely, as of now the system has difficulties in finding the removal location in 3D space when weeds are relatively large. This makes it difficult to estimate the root location of the weed, which should be the target for effective removal. 

During indoor tests, it has been shown that the efficiency of a single Spot system is still much lower than comparable wheeled systems. However, once full autonomy is achieved, we have justified that around 5 Spot robots are currently required to manage the weeds in a one-hectare field on a full-time basis during the pre-canopy closure period. We expect this to become possible in the future, especially if the system is further improved, making the agricultural sector more sustainable, and opening up possibilities for more targeted weed and crop management.

\section{Recommendations for future work}
\label{sec:recommendations}

This section provides several recommendations for enhancing the accuracy, efficiency, and autonomy of the proposed mobile robotic weed removal system. The suggestions focus on four key components: weed detection, workspace optimization, weed removal mechanics, and locomotion planner.

\subsection{Weed detection}
Significant improvements are required for outdoor weed detection, where two primary challenges must be addressed: enhancing the 2D RGB weed detection algorithm and improving the mapping from 2D RGB and depth pixel data to a precise 3D removal location. 

First, the current 2D RGB weed detection, utilizing the YOLO algorithm, demonstrated limited reliability in outdoor conditions due to insufficient data collected under a single lighting condition and weed growth stage. To enhance robustness, a diverse dataset containing various lighting conditions and weeds of different sizes should be gathered. This expansion will facilitate better generalization of the detection algorithm across different environmental conditions. Increasing the resolution of input data could also improve detection accuracy. Currently, images are processed at a resolution of \mbox{$320\times240$} pixels, where some weeds occupy only a few pixels, making it challenging even for human observers to differentiate them from other objects, such as leaves. Higher-resolution data could provide more details, enhancing the algorithm's ability to detect weeds accurately. This improvement requires higher computational power which can be addressed by upgrading the hardware capacity as was done in the architecture shown in \autoref{fig:integration-jetson}.

Furthermore, the method to determine the 3D weed removal location is, as mentioned before, one of the main bottlenecks of the current system. The current approach estimates the weed's center in the RGB data and assigns a depth to penetrate \SI{2}{\centi\meter} into the ground below the plant. For larger plants, this method is problematic, as the center of the plant canopy may not align with the root location, leading to inaccurate targeting by the milling bit. Utilizing depth data directly to identify the weed location could provide a more accurate 3D representation of the plant structure, improving the precision of the removal process. Another approach could be to train a network to directly detect the root location itself instead of the entire plant.

\subsection{Workspace optimization}
To use the full potential of the Spot robot, lower-level joint control is required instead of the high-level pose control applied in our work. This way, a custom-made pose controller can be made that increases the size of the feasible workspace $E$ to be much closer to the kinematically feasible workspace $K$. This not only offers advantages in increasing the workspace, but also enables finding the exact workspace when Spot is standing on uneven ground, as the constraints are exactly known. This allows for real-time adaptation of the workspace based on the robot's stance and even allows for intentionally positioning the robot in a specific way to maximize the available workspace.

\subsection{Weed removal}
Using the architecture in \autoref{fig:integration-jetson}, the weed removal controller has been improved by running all computations on board, enabling wired communication between the camera, the main computing unit, and the Spot robot, reducing the communication speed drastically. This also allows the adaptive target updating controller used for weed removal to run in a smooth manner because pose and camera readings can be time-synced without Spot having to stand stationary, which was not tested yet but would be interesting for future developments. 

\subsection{Locomotion planner}
To enable full autonomy, the simple locomotion planner in this work should be replaced by one that uses camera information for local navigation to account for, e.g., slight curves in paths, and uses some form of GPS for global navigation.

Additionally, the absence of an algorithm that avoids stepping on crops presents a challenge when working in narrower or intercropped fields. While the current method is sufficient for walking within rows without crops, more complex scenarios require an advanced obstacle avoidance system.

\section*{Acknowledgment}
This research was conducted as part of the Synergia consortium.
The Synergia project is organized and led by Wageningen University and Research in close cooperation with Next Food Collective as well as the Universities of Delft, Twente, Eindhoven, and Nijmegen. The authors have declared that no competing interests exist in the writing of this publication. Funding for this research was obtained from the Netherlands Organisation for Scientific Research (NWO grant 17626), IMEC-One Planet and other private parties.

The authors thank Brom Mechatronica for manufacturing the weeding tool and Ismail Elmasry for handling the electronics.

\appendix
\section{Inverse kinematics}
\label{app:inverse-kinematics}
Inverse kinematics (IK) are used to determine the joint angles required to position the body and legs of a robot in a desired configuration. This involves computing the body location, determining the initial positions of the four hips, and locating the feet relative to these frames.

\begin{figure}
    \centering
    \begin{subfigure}{\columnwidth}
        \centering

        \begin{tikzpicture}[scale=1.7,>=Stealth]

        \tikzset{
          joint/.style={circle,fill=black,inner sep=2pt},
          link/.style={thick},
          dashedlink/.style={thick,dashed},
          dottedlink/.style={thick,dotted},
          blueaux/.style={AccentBlue,dashed,thick},
          anglemark/.style={AccentBlue,thick}
        }
        
        \coordinate (J1) at (0,3.6);
        \coordinate (J2) at (2,4);
        \coordinate (J4) at (2.8,0);
        \coordinate (J3) at ($(J2)!0.55!(J4)$);

        \coordinate (J1x) at ($(J1)+(1,0)$);
        
        \coordinate (P) at (0,0);
        
        \draw[link] (J1) -- (J2);
        \draw[link] (J2) -- (J3);
        \draw[link] (J3) -- (J4);
        
        \node[joint,label=left:$J_1$] at (J1) {};
        \node[joint,label=above right:$J_2$] at (J2) {};
        \node[joint,label=left:$J_3$] at (J3) {};
        \node[joint,label=below:$J_4$] at (J4) {};
        
        \draw[dashedlink,<->] ($(J1)+(-0.04,0.2)$) -- ($(J2)+(-0.04,0.2)$)
          node[midway,above] {$l_1$};
        
        \draw[dashedlink,<->] ($(J2)+(0.2,0.04)$) -- ($(J4)+(0.2,0.04)$)
          node[midway,right] {$l_4$};
        
        \draw[dottedlink] (J1) -- ++(1.3,0);
        
        \draw[anglemark,black] ($(J2)+(-0.2,-0.04)$) -- ($(J2)+(-0.2,-0.04)+(0.04,-0.2)$) -- ($(J2)+(0.04,-0.2)$);

        \draw[anglemark]
          pic["\textcolor{black}{$\theta_1$}",draw=black,angle radius=16mm,angle eccentricity=1.2]
          {angle = J1x--J1--J2};
          
        \draw[anglemark]
          pic["$\alpha_1$",draw=AccentBlue,angle radius=8mm,angle eccentricity=1.3]
          {angle = J4--J1--J2};
        
        \draw[anglemark]
          pic["$\alpha_2$",draw=AccentBlue,angle radius=8mm,angle eccentricity=1.3]
          {angle = J3--J4--J1};

        \draw[blueaux,<->] (J1) -- (J4) node[midway,below left] {$A$};
        
        \draw[blueaux,<->] (J1) -- (P)
          node[midway,left] {$\Delta z$};
        
        \draw[blueaux,<->] (P) -- (J4)
          node[midway,below] {$\Delta y$};
        
        \draw[->] (-0.5,-0.3) -- (0.1,-0.3) node[right] {$y$};
        \draw[->] (-0.5,-0.3) -- (-0.5,0.3) node[above] {$z$};
        
        \end{tikzpicture}
        
        \caption{Front view of the leg schematics.}
        \label{fig:kinematics1}
    \end{subfigure}

    \vspace{4mm}

    \begin{subfigure}{\columnwidth}
        \centering

        \begin{tikzpicture}[scale=1.7,>=Stealth]

        \tikzset{
          joint/.style={circle,fill=black,inner sep=2pt},
          link/.style={thick},
          dashedlink/.style={thick,dashed},
          dottedlink/.style={thick,dotted},
          blueaux/.style={AccentBlue,dashed,thick},
          anglemark/.style={AccentBlue,thick}
        }
        
        \coordinate (J2) at (0.5,3.9);
        \coordinate (J3) at (-0.5,1.9);
        \coordinate (J4) at (1.0,0.4);
        \coordinate (Xproj) at ($(J2 |- J4) + (0,-0.2)$);

        \coordinate (J2x) at ($(J2) + (-1,0)$);
        \coordinate (J2y) at ($(J2) + (0,-1)$);

        \draw[link] (J2) -- (J3);
        \draw[link] (J3) -- (J4);
        
        \node[joint,label=above:$J_2$] at (J2) {};
        \node[joint,label=left:$J_3$]  at (J3) {};
        \node[joint,label=right:$J_4$] at (J4) {};
        
        \draw[dashedlink,<->] ($(J2)+({-0.1*0.1/sqrt(0.0125)},{0.05*0.1/sqrt(0.0125)})$) -- ($(J3)+({-0.1*0.1/sqrt(0.0125)},{0.05*0.1/sqrt(0.0125)})$)
         node[midway,left] {$l_2$};

        \draw[dashedlink,<->] ($(J3)+({-0.1*0.5*sqrt(2)},{-0.1*0.5*sqrt(2)})$) -- ($(J4)+({-0.1*0.5*sqrt(2)},{-0.1*0.5*sqrt(2)})$)
          node[midway,left] {$l_3$};
        
        \draw[dashedlink,<->] ($(J4)+(0.4,0)$) -- ($(J4 |- J2)+(0.4,0)$)
          node[midway,right] {$l_4$};

        \draw[dottedlink] (J2) -- ($(J2|-Xproj)+(0,-0.1)$);
        \draw[dashedlink,<->,AccentBlue] (Xproj) -- ($(J4 |- Xproj)$)
          node[midway,below,AccentBlue] {$\Delta x$};
        
        \draw[blueaux,<->] (J2) -- (J4) node[midway,right] {$B$};
        
        \draw[dottedlink] ($(J2)+(-0.6,0)$) -- ($(J4|-J2)+(0.5,0)$);
        \draw[anglemark,black]
          pic["$\theta_2$",draw=black,angle radius=8mm,angle eccentricity=1.3]
          {angle = J2x--J2--J3};
        
        \draw[anglemark,black]
          pic["$\theta_3$",draw=black,angle radius=7mm,angle eccentricity=1.3]
          {angle = J4--J3--J2};
        
        \draw[anglemark]
          pic["$\beta_2$",draw=AccentBlue,angle radius=9mm,angle eccentricity=1.3]
          {angle = J3--J2--J4};
        
        \draw[anglemark]
          pic["$\beta_1$",draw=AccentBlue,angle radius=17mm,angle eccentricity=1.2]
          {angle = J2y--J2--J4};
        
        \draw[->] (-1.0,0) -- (-0.3,0) node[right] {$x$};
        \draw[->] (-1.0,0) -- (-1.0,0.7) node[above] {$z'$};
        
        \end{tikzpicture}
        
        \caption{Side view of the leg schematics.}
        \label{fig:kinematics2}
    \end{subfigure}
    
    \caption{Schematic overview of the leg dimensions and auxiliary variables to solve the inverse kinematics.}
    \label{fig:kinematics} 
\end{figure}
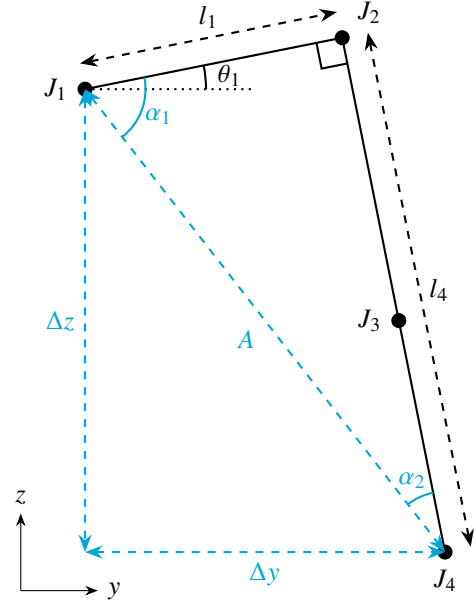
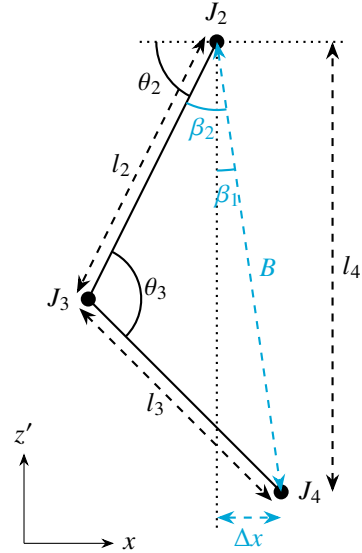

Each leg of the robot has four joints: $J_1, J_2, J_3$, and $J_4$. The joints $J_1$, $J_2$, and $J_3$ are responsible for the hip rotation $\theta_1$ about the $x$-axis, hip rotation $\theta_2$ about the $y'$-axis and knee rotation $\theta_3$ about the $y'$-axis, respectively. Joint $J_4$ is aligned with the foot and has no degrees of freedom. The $y'$-axis aligns with the hip link $l_1$. The $z'$-axis, as shown in \autoref{fig:kinematics2}, coplanar with joints $J_2$, $J_3$ and $J_4$, and perpendicular to the $x$-axis.

The IK are solved in the hip frame, which is perpendicular to the body frame located at $J_1$. Location $J_4$ is known in this frame to solve the IK. To derive the joint angles, the support variables $\alpha_1$, $\alpha_2$, $A$ and $l_4$ are defined as shown in \autoref{fig:kinematics1}, and the support variables $\beta_1$, $\beta_2$ and $B$ are defined as shown in \autoref{fig:kinematics2}. The joint angles $\theta_1$, $\theta_2$, and $\theta_3$ are found using:
\begin{align}
    A &= \sqrt{\Delta z^2 + \Delta y^2}, \\
    \alpha_2 &= \arcsin(l_1/A), \\
    \alpha_1 &= \pi - \alpha_2 - \frac{\pi}{2}, \\
    \theta_1 &= \alpha_1 - \arctan\!2(-\Delta z, \Delta y), \\
    l_4 & = \sqrt{\!\left(\Delta z + l_1\!\sin(\theta_1)\right)^2 + \left(\Delta y - l_1\! \cos(\theta_1) \right)^2}, \\
    B &= \sqrt{l_4^2+\Delta x^2}, \\
    \beta_1 & = \arctan\!2(\Delta x, l_4), \\
    \beta_2 & = \arccos\left(\frac{l_3^2-l_2^2-B^2}{-2l_2B}\right), \\
    \theta_2 &= -\beta_2 + \frac{\pi}{2} + \beta_1, \\
    \theta_3 &= \arccos\left(\frac{B^2-l_2^2-l_3^2}{-2l_2l_3}\right).
\end{align}

\section{Workspaces}
\label{app:workspace}
\begin{figure}
    \centering
    \begin{subfigure}{\columnwidth}
        \centering
        \includegraphics[width=\linewidth]{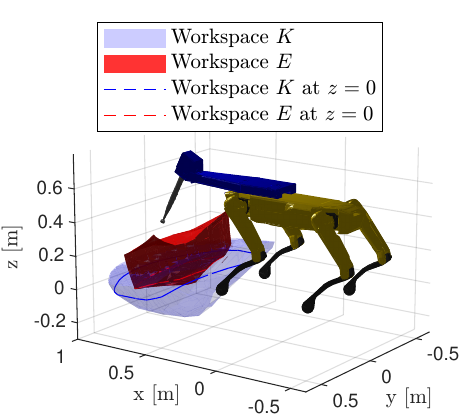}
        \caption{3D workspace of Spot with the custom weed removal tool. Although the workspace extends further along the positive $z$-axis, this extension is not depicted for clarity.}
        \label{fig:3D-workspace-1}
    \end{subfigure}

    \vspace{4mm}

    \begin{subfigure}{\columnwidth}
        \centering
        \includegraphics[width=\linewidth]{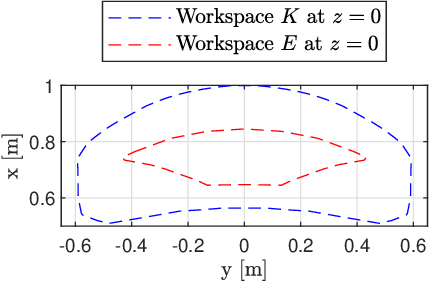}
        \caption{Cross-cut of the workspaces at $z = 0$.}
        \label{fig:2D-workspace-1}
    \end{subfigure}
    
    \caption{The workspaces for case 1.}
    \label{fig:workspace-case-1} 
\end{figure}

\begin{figure}
    \centering
    \begin{subfigure}{\columnwidth}
        \centering
        \includegraphics[width=\linewidth]{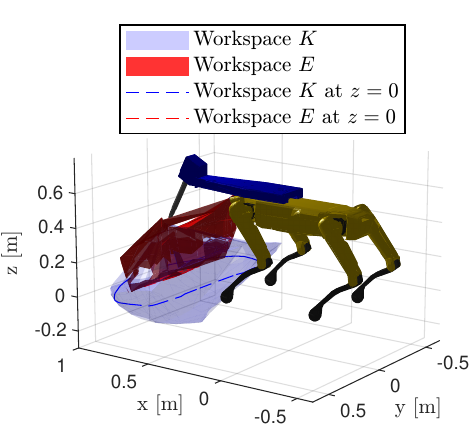}
        \caption{3D workspace of Spot with the custom weed removal tool. Although the workspace extends further along the positive $z$-axis, this extension is not depicted for clarity.}
        \label{fig:3D-workspace-2}
    \end{subfigure}

    \vspace{4mm}

    \begin{subfigure}{\columnwidth}
        \centering
        \includegraphics[width=\linewidth]{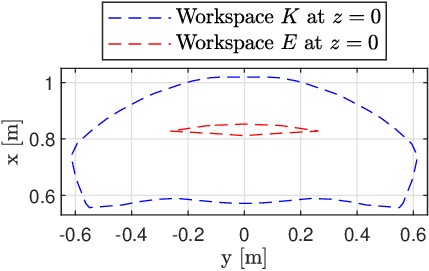}
        \caption{Cross-cut of the workspaces at $z = 0$.}
        \label{fig:2D-workspace-2}
    \end{subfigure}
    
    \caption{The workspaces for case 2.}
    \label{fig:workspace-case-2} 
\end{figure}

\begin{figure}
    \centering
    \begin{subfigure}{\columnwidth}
        \centering
        \includegraphics[width=\linewidth]{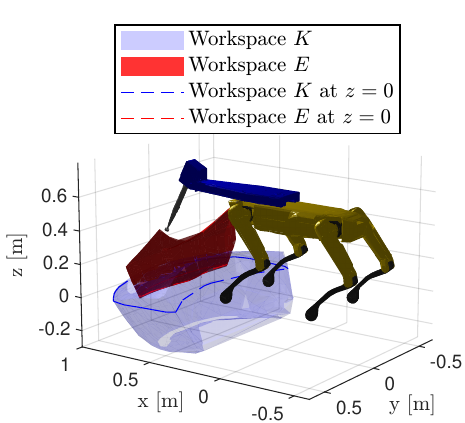}
        \caption{3D workspace of Spot with the custom weed removal tool. Although the workspace extends further along the positive $z$-axis, this extension is not depicted for clarity.}
        \label{fig:3D-workspace-3}
    \end{subfigure}

    \vspace{4mm}

    \begin{subfigure}{\columnwidth}
        \centering
        \includegraphics[width=\linewidth]{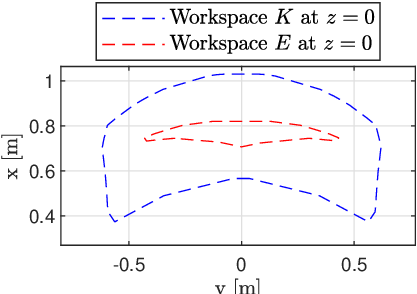}
        \caption{Cross-cut of the workspaces at $z = 0$.}
        \label{fig:2D-workspace-3}
    \end{subfigure}
    
    \caption{The workspaces for case 3.}
    \label{fig:workspace-case-3} 
\end{figure}

\begin{figure}
    \centering
    \begin{subfigure}{\columnwidth}
        \centering
        \includegraphics[width=\linewidth]{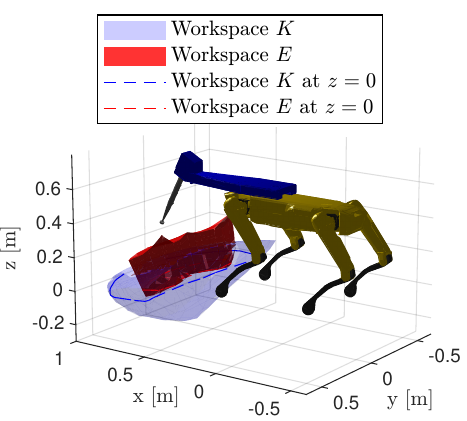}
        \caption{3D workspace of Spot with the custom weed removal tool. Although the workspace extends further along the positive $z$-axis, this extension is not depicted for clarity.}
        \label{fig:3D-workspace-4}
    \end{subfigure}

    \vspace{4mm}

    \begin{subfigure}{\columnwidth}
        \centering
        \includegraphics[width=\linewidth]{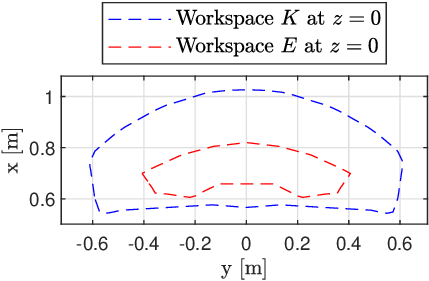}
        \caption{Cross-cut of the workspaces at $z = 0$.}
        \label{fig:2D-workspace-4}
    \end{subfigure}
    
    \caption{The workspaces for case 4.}
    \label{fig:workspace-case-4} 
\end{figure}

The workspaces for which the area is listed in \autoref{tab:reduced-workspace} are shown in \autoref{fig:workspace-case-1} to \autoref{fig:workspace-case-4} for cases 1 to 4, respectively.

\FloatBarrier

\bibliographystyle{plainnat}
\bibliography{References}

\end{document}